\documentclass[sw, crcready]{iosart2x}

\usepackage{eurosym}
\usepackage{graphicx}
\usepackage{amsmath}
\usepackage{amssymb}
\usepackage{caption}
\usepackage{mathrsfs} 
\usepackage{subcaption}
\usepackage{hyperref}
\usepackage{color}
\usepackage{algorithm, algorithmic}

\usepackage{multirow}
\usepackage{booktabs}
\usepackage{adjustbox}
\usepackage{array}
\usepackage{lscape}
\usepackage{rotating}

\usepackage{tikz}
\usetikzlibrary{patterns}
\definecolor{node}{HTML}{FF7900}
\definecolor{orangeborder}{HTML}{000000}
\definecolor{tablegrey}{HTML}{C0BCBC}
\usepackage{colortbl}
\definecolor{losange}{HTML}{000000}
\definecolor{carre}{HTML}{000000}
\definecolor{rond}{HTML}{000000}
\definecolor{orange}{HTML}{FF7900}

\pubyear{0000}
\volume{0}
\firstpage{1}
\lastpage{1}

\begin{document}

\begin{frontmatter}

\title{Uncertainty Management in the Construction of Knowledge Graphs: a Survey}
\runtitle{Uncertainty in the construction of knowledge graphs: A survey}

\begin{aug}
\author[A,B]{\inits{N.}\fnms{Lucas} \snm{Jarnac}\ead[label=e1]{lucas.jarnac@orange.com}%
\thanks{Corresponding author. \printead{e1}.}}
\author[A]{\inits{N.N.}\fnms{Yoan} \snm{Chabot}\ead[label=e2]{yoan.chabot@orange.com}}
\author[B,C]{\inits{N.-N.}\fnms{Miguel} \snm{Couceiro}\ead[label=e3]{miguel.couceiro@loria.fr}}
\address[A]{\orgname{Orange}, \cny{France}\printead[presep={\\}]{e1,e2}}
\address[B]{\orgname{Université de Lorraine, CNRS, LORIA}, Nancy, \cny{France}\printead[presep={\\}]{e3}}
\address[C]{\orgname{INESC-ID, Instituto Superior Técnico, Universidade de Lisboa}, Lisbon, \cny{Portugal}}
\end{aug}

\begin{abstract}
Knowledge Graphs (KGs) are a major asset for companies thanks to their great flexibility in data representation and their numerous applications, \textit{e.g.,} vocabulary sharing, Q/A or recommendation systems. 
To build a KG it is a common practice to rely on automatic methods for extracting knowledge from various heterogeneous sources. 
But in a noisy and uncertain world, knowledge may not be reliable and conflicts between data sources may occur.
Integrating unreliable data would directly impact the use of the KG, therefore such conflicts must be resolved.
This could be done manually by selecting the best data to integrate.
This first approach is highly accurate, but costly and time-consuming.
That is why recent efforts focus on automatic approaches, which represents a challenging task since it requires handling the uncertainty of extracted knowledge throughout its integration into the KG.
We survey state-of-the-art approaches in this direction and present constructions of both open and enterprise KGs and how their quality is maintained. 
We then describe different knowledge extraction methods, introducing additional uncertainty. 
We also discuss downstream tasks after knowledge acquisition, including KG completion using embedding models, knowledge alignment, and knowledge fusion in order to address the problem of knowledge uncertainty in KG construction.
We conclude with a discussion on the remaining challenges and perspectives when constructing a KG taking into account uncertainty.
\end{abstract}


\end{frontmatter}

\section{Introduction}
\label{section:introduction}

\begin{figure}[h]
     \centering
     \begin{subfigure}[b]{0.4\textwidth}
         \centering
         \includegraphics[width=\textwidth]{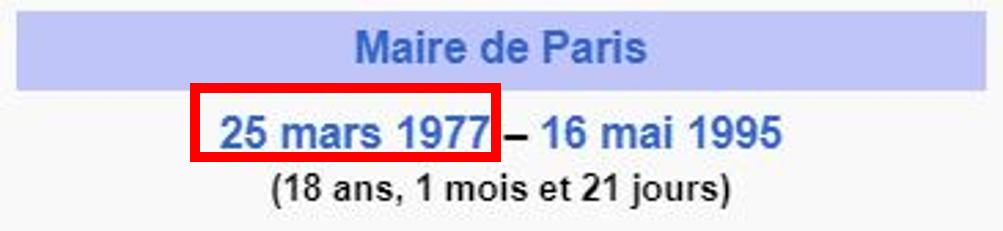}
         \caption{French Wikipedia page of Jacques Chirac.}
     \end{subfigure}
     \hfill
     \begin{subfigure}[b]{0.4\textwidth}
         \centering
         \includegraphics[width=\textwidth]{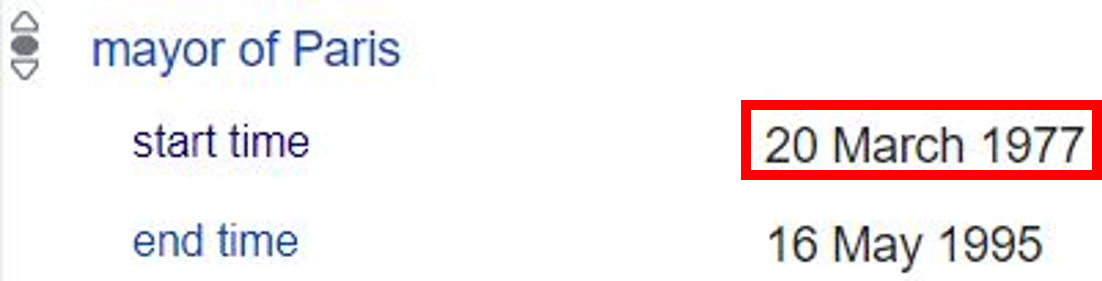}
         \caption{Wikidata page of Jacques Chirac.}
     \end{subfigure}
        \caption{Illustration of a contradiction between two sources. One of the sources claims that the mandate of Jacques Chirac as mayor of Paris began on March 25 while the other claims that it began on March 20.}
        \label{fig:jacques-chirac}
\end{figure}

Huge amounts of data expressed in the form of tables, texts, or databases are generated by organizations every day.
When using these data within an organization, we have to deal with uncertainty, as the data often suffer from contradictions and differences in specificity.
These are the effect of incompleteness, vagueness, fuzziness, invalidity, ambiguity, and timeliness leading to uncertainty about the correctness of the data~\cite{djebri2022}.
The uncertainty can be due to the source of the data (\textit{e.g.,} a document written by an expert \textit{vs.} a non-expert in the field concerned) or in the data itself (\textit{e.g.,} a scientific supposition where the fact is not yet well-defined but accepted by consensus as it is).
For example, on the French Wikipedia page of the former president of France Jacques Chirac\footnote{\url{https://fr.wikipedia.org/wiki/Jacques_Chirac}}, we can read that he was the mayor of Paris from March 25, 1977 to May 16, 1995 while on Wikidata\footnote{\url{https://www.wikidata.org/wiki/Q2105}} it is mentioned that he was the mayor of Paris from March 20, 1977 to May 16, 1995 as depicted in Figure~\ref{fig:jacques-chirac}.
In addition, data are not settled over time~\cite{reynolds2014}.
Some facts are known to change including all facts that involve a period of time for which the fact is valid (\textit{e.g.,} the mandate of a president or the place of residence of a person) or knowledge in specific domains may be known to change regularly, which is often the case for History where excavations can modify knowledge which, until today, was considered established.

Many knowledge graphs (KGs) have been built to represent such data in recent years, and they
have become major asset for organizations, since KGs can support various downstream tasks such as knowledge and vocabulary indexation, as well as other applications in recommendation systems, question/answering systems, knowledge management, or search engine systems~\cite{hogan2022, Noy2019}.
To build or enrich a KG and reconcile uncertain data, we can rely on manual approaches (\textit{e.g.,} domain experts) but this is a time-consuming and tedious process.
Alternatively, it is common to leverage automatic knowledge extraction approaches that handle large volumes of data from various heterogeneous sources, \textit{e.g.,} texts~\cite{Dong2014}, tables~\cite{liuChabot2023}, or databases to ensure the coverage of the KG.
These automatic approaches are usually based on three main steps: 
\begin{enumerate}
    \item Extraction of knowledge from documents.
    \item Detection of duplicates and conflicts between extracted knowledge. Conflicts occur from differences of specificity or knowledge contradictions.
    \item Fusion of aligned knowledge: once detection is completed, conflicting knowledge should be reconciled.
\end{enumerate}
However, each of the aforementioned steps is error-prone and increases the uncertainty on extracted knowledge due to the performance of the algorithms~\cite{jungKim2019, wick2013, LiLiCATD2014}.

Uncertainty may also be found in knowledge, which we can distinguish two types: {\it objective knowledge} in which a single value is accepted (\textit{e.g.,} the mandate of Jacques Chirac where only one period of time is the true value), and {\it subjective knowledge} in which several values can be accepted according to their context and point of view (\textit{e.g.,} the number of participants in a protest depending on the counting technique). 
Most KG construction methods do not take into account noisy facts and the uncertainty inherent in extraction algorithms and knowledge, which may impact downstream applications.
Therefore, there is a need to reconcile knowledge units extracted from heterogeneous sources before integrating them into the KG in order to obtain a single or multiple representations that are as reliable and fair as possible~\cite{Noy2019}.
In this survey, we review different approaches to integrate knowledge uncertainty in the main steps of KG construction with up-to-date knowledge fusion methods and its representation in the graph.
\cite{Paulheim2017} surveys approaches and evaluation methods for KG refinement, particularly KG completion and error detection methods.
In~\cite{LiGao2015}, the authors review truth discovery methods used in knowledge fusion before 2015.
However, to the best of our knowledge, there is no survey about uncertainty handling in the construction of KGs.

The remainder of this survey is structured as follows.
In Section~\ref{section:research-methodology}, we describe our research methodology which led us to write this survey.
We introduce the definition of KGs with some well-known KGs that have been built in recent years, then tools and quality metrics considered in KGs construction in Section~\ref{section:knowledge-graphs}.
We present some knowledge extraction approaches and why they lead to uncertain knowledge in Section~\ref{section:kg-construction}.
An ideal knowledge integration pipeline handling the uncertainty of knowledge is provided in Section~\ref{section:kg-refinement}, while the steps of knowledge refinement from the pipeline are described in Section~\ref{section:uncertainty-consideration} for uncertainty consideration in KG representation learning, in Section~\ref{section:kg-alignment} for knowledge alignment, and in Section~\ref{section:kg-fusion} for knowledge fusion. 
The solutions for uncertainty representation in KGs are then listed and depicted in Section~\ref{section:uncertainty-representation}.
We also discuss some perspectives on the use of uncertainty in the KG ecosystem in Section~\ref{section:perspectives}, before concluding this survey in Section~\ref{section:conclusion}.

\section{Research Methodology}
\label{section:research-methodology}

This paper aims at surveying methods to construct a KG from uncertain knowledge.
In this section, we provide our research methodology for discovering and selecting papers on this purpose.
To find papers of interest, we mainly used the Google Scholar search engine and created alerts with the following keywords: KG fusion, multi-source knowledge fusion, KG resolution, KG quality, knowledge fusion, KG reconciliation, KG alignment, KG matching, KG resolution, KG cleaning.
The aforementioned keywords have been combined with the keyword ``uncertain'' and the terms ``knowledge'', ``data'', and ``information'' used interchangeably, \textit{e.g.,} ``uncertain knowledge fusion'', ``uncertain data fusion'', or ``uncertain information fusion''.

\begin{figure*}[h]
	\centering
	\includegraphics[scale=0.55]{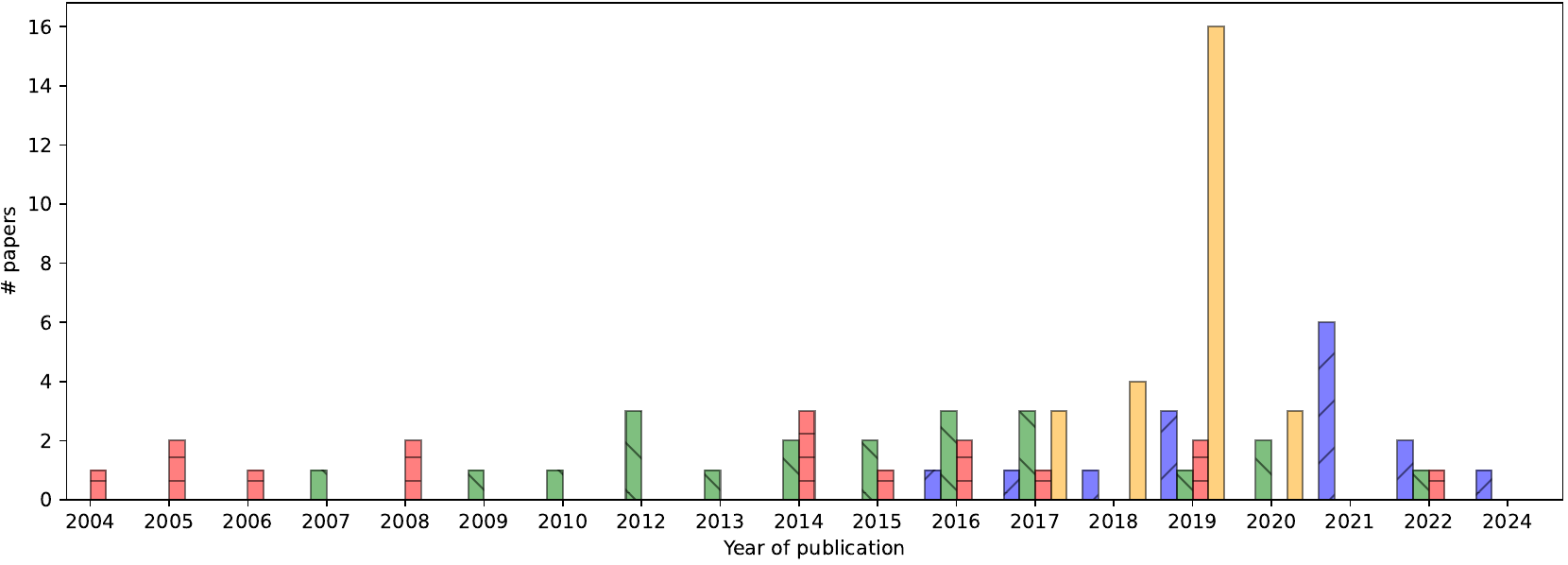}
	\caption[]{Distribution of paper publication years according to uncertain KG embedding (\raisebox{-1pt}{\begin{tikzpicture}[scale=0.5]
        \draw[color=blue, fill=blue, opacity=0.5] (0,0) rectangle (0.25,0.8);
        \draw[pattern=north east lines=5pt, pattern color=black, opacity=0.3] (0,0) rectangle (0.25,0.8);
    \end{tikzpicture}}), knowledge fusion (\raisebox{-1pt}{\begin{tikzpicture}[scale=0.5]
        \draw[color=green, fill=green, opacity=0.5] (0,0) rectangle (0.25,0.8);
        \draw[pattern=north west lines=5pt, pattern color=black, opacity=0.3] (0,0) rectangle (0.25,0.8);
    \end{tikzpicture}}), knowledge alignment (\raisebox{-1pt}{\begin{tikzpicture}[scale=0.5]
        \draw[color=orange, fill=orange, opacity=0.5] (0,0) rectangle (0.25,0.8);
    \end{tikzpicture}}), uncertainty representation (\raisebox{-1pt}{\begin{tikzpicture}[scale=0.5]
        \draw[color=red, fill=red, opacity=0.5] (0,0) rectangle (0.25,0.8);
        \draw[pattern=horizontal lines=5pt, pattern color=black, opacity=0.3] (0,0) rectangle (0.25,0.8);
    \end{tikzpicture}}).}
	\label{fig:bar-publication-year}
\end{figure*}

For the selection of the papers, we proceeded as follows:
(1) we look at the title of the paper, if it is relevant and seems to be related to one of the topics we were looking for, then we read the abstract;
(2) if the paper presents a method related to uncertain KG construction or a method that is not related to KGs but which could be extrapolated to a KG, we selected it.
We provide Figure~\ref{fig:bar-publication-year} that depicts the distribution of paper publication years according to four applications of KG refinement: uncertain embedding, knowledge fusion, knowledge alignment, and uncertainty representation.
We observe that the representation of uncertainty has been addressed in the literature since 2004, while the vector space representation of uncertainty is a more recent research topic (since 2016).
The number of papers on knowledge fusion mentioned in this survey has remained constant over the last few years.
This is due to new models based on deep learning that are being explored to tackle these tasks.

\section{Knowledge Graphs}
\label{section:knowledge-graphs}

Before going into further detail on reconciliation approaches, it is important to define KGs, which are the core of this survey.
KGs provide a structural representation of knowledge that is captured by the relations between entities in the graph.
The KGs provide a concise and intuitive data representation and abstraction, making them an ideal tool to manage knowledge of organizations in a sustainable way or to support search and querying applications~\cite{hogan2021}, which led several companies to build their own KGs~\cite{Noy2019}.
 
In this section, we provide a definition of KG, and we describe some well-known KGs including open KGs and Enterprise Knowledge Graphs (EKGs) and how their consistency is maintained in Section~\ref{subsection:kg-definition}.

\subsection{What is a Knowledge Graph?}
\label{subsection:kg-definition}

What is the meaning of knowledge?
Data are uninterpretable signals (\textit{e.g.,} numbers or characters). 
Information is data equipped with a meaning.
In~\cite{schreiber2000}, the authors define knowledge as data and information that enter into a generative process supporting tasks and creating new information.
A knowledge graph (KG) is ``a graph of data intended to accumulate and convey knowledge of the real world, whose nodes represent entities of interest and whose edges represent potentially different relations between these entities''~\cite{hogan2021}.

\begin{figure}[h]
	\centering
	\includegraphics[scale=0.75]{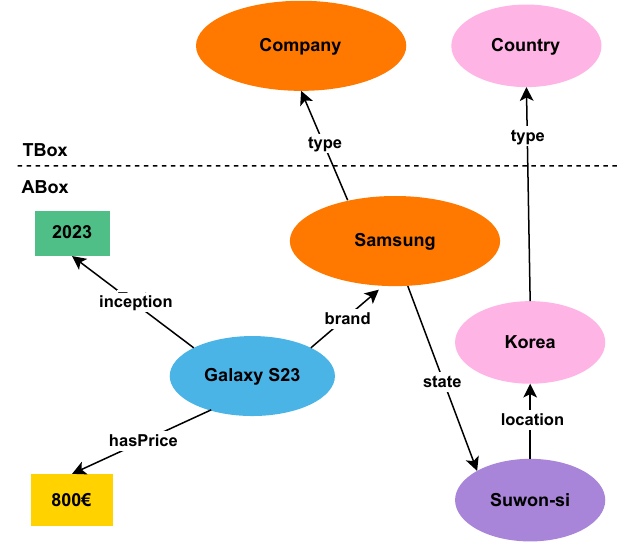}
	\caption{TBox stands for terminology box that contains classes and properties; ABox stands for assertion box that contains instances (\textit{e.g.,} Galaxy S23) and values (\textit{e.g.,} ``2023'').}
	\label{fig:abox-tbox}
\end{figure}

Formally, KGs are directed and labeled multigraphs $(\mathcal{E},\mathcal{R},\mathcal{T})$, where $\mathcal{E}$ is the set of entities, $\mathcal{R}$ is the set of relations, and $\mathcal{T}$ is the set of triples $\left\langle \text{subject, predicate, object} \right\rangle$ that are the atomic elements of KGs, also called facts.
The subject and object are represented by nodes, the predicate indicates the nature of the relationship holding between the subject and object represented by an edge in the KG~\cite{hogan2021,Noy2019}.
For instance, such a triple could be $\left\langle \text{Suwon-si, location, Korea} \right\rangle$ as illustrated in Figure~\ref{fig:abox-tbox}.
The classes and relationships of entities are defined through a schema or otherwise named an ontology, which can be itself represented as a graph embedded in the KG~\cite{ehrlinger2016, Dong2023}.

In a KG, two boxes coexist, namely Terminology Box (TBox) and Assertion Box (ABox).
TBox defines classes and properties and ABox contains instances of classes defined in the TBox.
For example, in Figure~\ref{fig:abox-tbox}, ``Company'' and ``Country'' are concepts defined by the ontology, ``Galaxy S23'', ``Samsung'', and ``Korea'' are the instances of these concepts while ``2023'', ``800\euro'' are literals \textit{i.e.,} attributes that characterize an entity.
This data representation in semi-structured form, defined by its ontology, offers a clear and flexible semantic representation whose classes and relations can easily be added and connect a large number of domains~\cite{Dong2023}.

\subsection{Open Knowledge Graphs}
\label{subsubsection:open-kgs}

For the last few years, some KG construction projects that link general knowledge about the world have appeared.
The best known is probably \textbf{Wikidata}\footnote{\url{https://www.wikidata.org/wiki/Wikidata:Main_Page}}, a large, free, and collaborative KG supported by the Wikimedia Foundation.
It is a multilingual general-purpose KG that contains more than 100 million elements~\cite{vrandecic2014}.
The structure of Wikidata is based on property-value pairs, where each property and entity is an element.
A typical entity also contains labels, aliases, descriptions, links and mentions to Wikipedia articles.
To maintain the quality of data, some constraints such as properties or unique values constraints alert the user in a case of suspicion of the input content (\textit{e.g.,} constraint violations).
Wikidata also keeps the sources and references of provenance of entities to ensure their traceability, which is one of the requirements for KG quality~\cite{Weikum2021}.

\textbf{NELL} is a research project led by Carnegie Mellon University in which an intelligent computer agent that ran continuously between January 2010 and September 2018 according to the official NELL website\footnote{\url{http://rtw.ml.cmu.edu/rtw/}} and that every day extracted knowledge from texts, tables, and lists from the web to feed a Knowledge Base (KB)~\cite{carlson2010}.
To maintain the consistency of the KB, the knowledge integrator of NELL exploits relationships between predicates by respecting mutual exclusion and type checking information.
On top of that, NELL components provide a probability for each candidate and a summary of the source supporting it, hence which can be qualified as a probabilistic KB.

\textbf{YAGO} is an ontology built on statements of Wikipedia that combines high coverage with high quality~\cite{suchanek2007}.
The data model of YAGO is based on entities and binary relations extracted from WordNet and Wikipedia.
A manual evaluation is performed to verify the quality of data.
To do this, facts are randomly selected with their respective Wikipedia pages that are used as Ground Truth (GT).

\textbf{DBpedia} is a multilingual knowledge base built by extracting structured information from Wikipedia (\textit{e.g.,} infoboxes) and makes this information accessible on the Web~\cite{bizer2009}.
Since DBpedia is populated from Wikipedia pages in different languages, data retrieved are sometimes conflicting.
To manage these conflicts, DBpedia has a module called \textit{Sieve} which performs a quality assessment by computing some metrics such as the ``recency'' or the ``reputation'' of data before applying a fusion step based on these dimensions~\cite{mendes2012, bryl2014}.

\textbf{Freebase} is a graph created in 2007 which provides general human knowledge and which aims to be a public directory of world knowledge.
A component included in Freebase called Mass Typer, allows a user to complete data and reconcile it semi-automatically with data already present in Freebase by performing three actions: merge, skip, or add the data.
Then acquired by Google and used to support systems like Google Search, Google Maps, and Google KG, nowadays, Freebase is closed, and its knowledge has been transferred to Wikidata~\cite{Bollacker2008}.

\textbf{ConceptNet} is the KG version of the Open Mind Common Sense project that contains information about words from several languages and their roles in natural language.
It was built by collecting knowledge from multiple data sources namely Open Mind Common Sense, Wiktionary, games with a purpose for harvesting common knowledge, Open Multilingual WordNet, JMDict, OpenCyc, and DBPedia.
Each node corresponds to a word or a sentence, and the relations between nodes are attached to numerical values that intend to represent the level of uncertainty about the relation~\cite{Speer2016}.

\subsection{Enterprise Knowledge Graphs}\label{subsubsection:enterprise-kgs}

EKGs are major assets for companies since they can support various downstream applications including knowledge/vocabulary sharing and reuse, data integration, information system unification, search, or question answering~\cite{galkin2016,Noy2019,Sequeda2021}.
This led companies such as Google, Microsoft, Amazon, Facebook, Orange and IBM, to build their own KGs~\cite{Noy2019, jarnac2022}.

For instance, Microsoft built {\bf Bing KG} to answer any kind of question through Bing search engine.
With a size of about two billion entities and 55 billion facts according to~\cite{Noy2019}, it contains general information about the world like people, or places and allows users to take actions like watching a video or buying a song.
Alternatively, KGs can also increase understanding of user behavior.

It is the case of the {\bf Facebook KG} that establishes links between the users as well as interests of users, \textit{e.g.,} movies or music tastes.
The Facebook KG is the largest social graph with about 50 million entities and 500 million statements in 2019.
To handle conflicting information, the Facebook KG removes information if the associated confidence is low, otherwise, conflicting information is integrated with its provenance and the estimated confidence of the information.

{\bf Yahoo KG}~\cite{torzek2018} offers different services such as a search engine, a discovery system to relate entities, or for entity recognition in queries and text.
To build their KG, they leverage Wikipedia and Freebase as the backbone of the KG and use various complementary data sources to maximize the relevance, comprehensiveness, correctness, freshness, and consistency of knowledge.
They mainly validate the data \textit{w.r.t.} the ontology and through a user interface that enables entities to be corrected and updated.

Also, Orange bootstraps its KG from a set of terms of interest from a enterprise repository~\cite{jarnac2022}.
These terms of interest are aligned with equivalent Wikidata entities before applying an expansion to retrieve the neighborhood to extend their KG.
To ensure the quality of this initial KG, pruning methods based on Euclidean distance in the embedding space, degrees of Wikidata entities, or a method based on analogical inference are used~\cite{jarnac2022, jarnacCouceiro2023}.

In~\cite{Noy2019}, the authors mention future challenges including disambiguation, knowledge extraction from unstructured and heterogeneous sources, and knowledge evolution management in the process of KG construction.
We discuss some of these challenges in the next section.

\section{Knowledge Acquisition}
\label{section:kg-construction}

The previous section introduced some KGs including open KGs and EKGs.
To build such KGs, we could rely on knowledge extraction that is the first step of knowledge integration process.
In this section, we present what knowledge extraction is and some well-known automatic approaches that contribute to uncertainty in Section~\ref{subsection:extraction} that extract knowledge from: texts (Section~\ref{subsubsection:texts}), Web (Section~\ref{subsubsection:web}), and Large Language Models (LLMs) with the recent interest in probing methods (Section~\ref{subsubsection:probing}).
Finally, the definition of KG quality and related metrics are provided in Section~\ref{subsection:metrics-quality}.

\subsection{Knowledge Extraction}
\label{subsection:extraction}

Methods to populate a KG depend on the knowledge domain and the desired graph coverage. 
For example, a first method could rely on the knowledge of domain experts and populate the graph manually (\textit{e.g.,} by crowdsourcing such as Wikidata~\cite{vrandecic2014} or Freebase~\cite{Bollacker2008}). 
However, this is a time-consuming process, particularly if the graph is intended to be large and it can suffer from quality issues~\cite{brasileiro2016, shenoy2022}. 
Furthermore, such open KGs have a large community that enterprises or specific KGs may not have.
Therefore, large KGs such as Google, Amazon and Bing KGs rely on automatic construction methods~\cite{Noy2019}.
In the following sections, we present the different tasks involved in knowledge extraction from different types of data sources.

\subsubsection{From texts}
\label{subsubsection:texts}

\begin{figure}[h]
	\centering
	\includegraphics[scale=1]{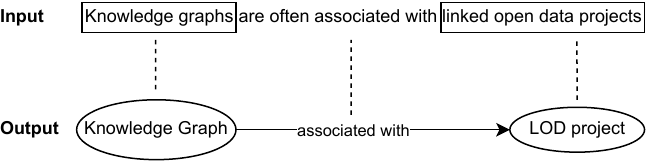}
	\caption{Illustration of knowledge extraction from a single sentence. LOD stands for Linked Open Data.}
	\label{fig:knowledge_extraction}
\end{figure}

For a long time, the majority of data was represented and exchanged in the form of text~\cite{martinez-rodriguez2018, pan2023}. 
Texts in all their forms (e.g. reports, articles, or any other textual document) are an invaluable source of information, as they are the most widely used data formats in the world (\textit{e.g.,} the scientific research area, where knowledge is communicated via scientific articles~\cite{jaradeh2019}).
To leverage knowledge from texts as data sources to enrich a KG, we rely on the task called Information Extraction (IE) (or knowledge acquisition).
IE transforms unstructured information in text form into structured information, \textit{i.e.,} $(entity_{1}, relation, entity_{2})$ triples~\cite{niklaus2018}.
The aim of information extraction is to identify entities, their attributes, and their relationships with other entities in text~\cite{wu2019}. 
In general, this task is separated into several sub-tasks: Entity Recognition (ER) or Named Entity Recognition (NER) and Relation Extraction (RE).
Figure~\ref{fig:knowledge_extraction} depicts the input and output of a text-based knowledge extraction task.
NER aims to identify named entities into the text and classify them to general types, while RE extracts semantic relationships that occur at least between two entities~\cite{zhong2023, harnoune2021}.

There are two main IE approaches in the literature~\cite{niklaus2018}:
Traditional Information Extraction (Traditional IE) and Open Information Extraction (Open IE).
Traditional IE relies on manually defined extraction patterns or patterns learned from manually labeled training examples~\cite{niklaus2018}. 
However, if the domain of interest evolves, the user must redefine the extraction patterns.	
Open IE does not rely on predefined patterns and faces three challenges~\cite{yates2007, niklaus2018}: automation, text heterogeneity and scalability. 
Automation means that the information extraction system must rely on unsupervised extraction strategies.
Heterogeneity stands for the difference of text types across knowledge, \textit{e.g.,} a scholar journal versus a popular science journal.
Because the extractions are performed in an unsupervised manner \textit{i.e.,} without any labelling or predefined schema to support the extraction, it implies higher uncertainty in the extracted knowledge.
Furthermore, as text types are heterogeneous due to its unstructured form, knowledge extraction patterns are more general and can lead to different levels of specificity of knowledge.
Finally, the system must be able to handle large volumes of text for scalability reasons.
The most widespread approaches to tackling these issues consist of pipelines composed of methods based on Natural Language Processing (NLP)~\cite{wu2019}.

One of the earliest examples of a Traditional IE systems is \textbf{KnowItAll}~\cite{etzioni2004} that automates the domain-independent extraction of large collections of facts (\textit{i.e.,} triples) from the Web. 
It is nevertheless supported by an extensible ontology and a minimal set of generic rules to extract entities and relations contained in its ontology.
KnowItAll includes four components: an extractor, a search system, an evaluator, and a database. 
Its extractor instantiates a set of extraction rules for each class and relation based on a generic domain-independent pattern, for example $\textit{``cities such ...''} \to \textit{``cities such as Paris, Stockholm, ...''}$ deduces that Paris and Stockholm are instances of a ``City'' class. 
The search component, which includes 12 search engines such as Google, applies queries based on the extraction rules, \textit{i.e.,} \textit{``Cities such as''}, then retrieves the web pages and applies the extractor.  
An evaluator leverages the statistics provided by search engines to assess the probability that the extracted relationships are valid. 
Once the extracted data has passed through these three components, it is stored into a relational database.

Traditional approaches for information extraction rely on an extractor for each target relation based on labeled training examples (\textit{e.g.,} pre-designed extraction patterns). 
However, these approaches do not address the problem of extraction on large corpora whose relations are not all specified in advance~\cite{fader2011} whereas Open IE no longer relies on predefined patterns and allows new information to be explored~\cite{niklaus2018}. 

For example, \textbf{TextRunner}~\cite{yates2007} that introduced the concept of Open IE, extracts a set of relational tuples without human input required. 
TextRunner is described by three components. 
The first one is a single-pass extractor that labels the text with part-of-speech tags (PoS) (\textit{i.e.,} grammatical tagging) and extracts $(e_{1}, r, e_{2})$ triples. 
The second component is a self-supervised classifier trained to detect the correctness of the extraction. 
Finally, the last component is a synonym resolver that groups together synonymous entities and relations, since TextRunner has no predefined relations on which extractions are guided.

A slightly more recent approach is \textbf{ReVerb}~\cite{fader2011}. 
Using constraints, this method aims to resolve the inconsistent extractions of previous Open IE models due to predicates composed of a verb and a noun. 
Two types of constraints are introduced on relational sentences: a syntactic constraint and a lexical constraint. 
Firstly, the syntactic constraint imposes the relational sentence to start either with a verb, a verb followed by a noun, or a verb followed by nouns, adjectives, or adverbs. 
Regarding the lexical constraint, it focuses on relations that can take many arguments and not on very specific relations. 
According to the results, these additional constraints allow ReVerb to outperform TextRunner. 
In addition, ReVerb assigns a confidence score to extractions from a sentence by applying logistic regression classification. 
To do this, extractions of the form $(x, r, y)$ from a sentence $s$ and for 1000 sentences were labeled as valid or invalid, and 19 features such as ``$s$ begins with $x$'', ``$x$ is a proper noun'', ``$(x, r, y)$ covers all words in $s$'' were used as input variables for the logistic regression model.
Such confidence scores can be used for downstream knowledge extraction tasks to support their integration into the KG (see Section~\ref{subsection:requirements}).

\textbf{OLLIE}~\cite{mausam2012} expands the syntactic scope of relations phrases to cover much larger number of expressions and allows additional context information such as attribution and clausal modifiers. 
The authors argue that other models lack context on extracted relations. 
Hence, compared to previous methods, OLLIE introduces a new component that analyzes the context of an extraction when the extracted relation is not mentioned as factual in the text. 
This context is attached to each extracted relation and models the validity of the information expressed (\textit{e.g.,} mentions of ``$\textit{according to}$'' in a sentence).
In~\cite{jaradeh2023}, the authors present multiple components involved in different IE pipelines in the literature. 
They propose several combinations of these components and evaluate them in a complete pipeline that includes four steps in the PLUMBER framework: Coreference Resolution, Triples Extraction, Entity Linking and Relation Linking.
40 reusable search components are combined, representing 432 distinct information extraction pipelines. 
Further information is provided in~\cite{jaradeh2023}.

\subsubsection{From the Web}
\label{subsubsection:web}

The Web contains an huge amount of data. 
It is probably the most widely used tool for exchanging knowledge between people (\textit{e.g.,} in the form of HTML texts). 
Therefore, it represents an invaluable data source for building KGs.
However, the latter suffers from uncertain facts, in part due to the fact that anyone can edit it.
In this context, it is necessary to select reliable data sources from the Web and to implement approaches for assessing the reliability of the extracted knowledge.
In this section, we present some KGs that have been built from the Web.

\textbf{NELL}~\cite{mitchell2018} has been extracting facts from the Web continuously since January 2010, and aims to improve over time. 
NELL is a system that takes an initial ontology as input and reads facts from the Web and removes the incorrect ones from a set of labeled data and user feedback on the trustworthiness of the extracted facts.
The core of NELL consists of learning thousands of tasks to classify extracted noun phrases into categories, to find the confident relations for each pair of noun phrases, and to identify synonymous noun phrases.
Then, the extracted facts are stored in a KB with their provenance and confidence score computed during the relation classification step.

\textbf{Knowledge Vault}~\cite{Dong2014} is a probabilistic KB that combines extractions from Web content and prior knowledge derived from existing knowledge repositories such as Freebase.
They rely on the Local Closed World Assumption, \textit{i.e.,} for a set of existing object values $O(s,p)$ from an existing KG that contains a set of $(s,p,o)$ triples, a candidate triple $(s,p,o)$ is correct if $(s,p,o) \in O(s,p)$. 
However, if $(s,p,o) \notin O(s,p)$ and $\left| O(s,p) \right|> 0$, the triple is incorrect.
Hence, this assumption can be difficult to adopt in the construction of an EKG.
To merge the extractors (four different fact extraction methods: text documents, HTML trees, HTML tables, and Human Annotated pages) they define a feature vector $f$ for each extracted triple, then apply a binary classifier to compute the probability that the fact is true.
They assume that the confidence scores from each extractor are not necessarily on the same scale.
Therefore, to cope with this issue, they apply a Platt scaling method that fits a logistic regression model to the confidence scores in order to obtain a probability distribution. 
Concerning the fusion task, they construct a feature vector $f$ for each extracted triple and apply a binary classifier to compute the probability of the fact to be true given the feature vector.
Each predicate is associated to a distinct classifier.
Each feature vector contains the square root of the number of sources where the extractor extracted this triple and the mean score of the extractions from this extractor.

\textbf{Probase}~\cite{wu2012} does not consider knowledge extracted from the Web to be deterministic, but models it using probabilities. 
The authors argue that existing KBs and taxonomy construction methods do not have sufficient concept coverage for a machine to understand the text in natural language. 
Probase includes the uncertainties of the extracted knowledge (specifically vagueness and inconsistencies that are due to the knowledge and to flawed construction methods). 
It was built from 1.6 billion web pages from an iterative learning algorithm that extracts pairs $(x, y)$ that verify an \textit{isA} relation between $x$ and $y$, then a taxonomy construction algorithm organizes these extracted pairs into a hierarchy. 
In Probase, facts have probabilities that measure their plausibility and typicality. 
Plausibility is computed from multiple features \textit{e.g.,} the PageRank score, the patterns used to extract \textit{isA} pairs, or the number of sentences where \textit{x} or \textit{y} is present with its respective role (sub or super concept).
Typicality is then computed as a function of plausibility and the number of evidences of the fact, \textit{i.e.,} the number of sentences in which the fact is mentioned.

\subsubsection{Probing}
\label{subsubsection:probing}

With the arrival of Deep Learning (DL) models and Large Language Models (LLMs), some triple extraction tasks are now successfully carried out by such models. \cite{nayak2021} reviews some of them such as Graph-Based Neural Models, CNN-based model, Attention-Based Neural model and others applied to a specific knowledge domain.
Also, with significant advances in LLMs and the fact that they are trained on a wide variety of information sources, some researchers have shifted their attention to  KG construction by leveraging the knowledge learned by LLMs~\cite{pan2023}.
For example, a workshop on KB construction from pre-trained language models (KBC-LM\footnote{\url{https://lm-kbc.github.io/workshop2024/}}) and a challenge on language models for KB construction (LM-KBC\footnote{\url{https://lm-kbc.github.io/challenge2024/}}) are now proposed at the International Semantic Web Conference (ISWC).

In~\cite{harnoune2021} the authors use the BERT model for NER and RE tasks to build a biomedical KG.
In~\cite{hao2023}, the authors exploit knowledge encoded in LLM parameters (\textit{a.k.a.} parametric knowledge~\cite{pan2023}) to feed a KG by harvesting knowledge for relations of interest. 
To illustrate their method, they provide an example of knowledge extraction for the ``\textit{potential\_risk}'' relation. 
The input contains a prompt such as ``The potential risk of A is B'' with a few shot of seed entity pairs that validate the relationship, \textit{e.g.,} $\textit{(eating candy, tooth decay)}$. 
Then, the entity pairs obtained at the output of the LLM are ranked according to a consistency score computed \textit{w.r.t.} the compatibility scores between entity pairs.
From such point of view, knowledge can be directly extracted from models without visible difficulty by performing aggregation of conflicting information.

However, Pan \textit{et al.}~\cite{pan2023}  explore possible interactions and synergies between KGs and LLMs including the construction of KGs from LLMs and raise several issues.
LLMs can be used to extract knowledge directly, but this mainly applies to generic, non-specific domains and perform poorly on specific domains.
They also lack accuracy in numerical facts such as the birthdate of a person and knowledge of long-tail entities, or have difficulty to memorize them.
In addition, LLMs are subject to various biases (\textit{e.g.,} gender bias) that are inherent to training data.
Finally, LLMs do not provide any provenance or reliability information of extracted knowledge~\cite{pan2023}, which can be an obstacle for many knowledge fusion approaches presented in Section~\ref{section:kg-fusion}.
In~\cite{zhu2023}, the authors evaluate the ability of LLMs, in particular GPT-3.5, ChatGPT and GPT-4, on KG construction and reasoning (\textit{i.e.,} link prediction and question answering) tasks under different settings, namely zero-shot and one-shot. 
The authors also point out LLMs fail to outperform state-of-the-art models for KG construction and have limitations to recognize long-tail knowledge.

\subsection{Quality and metrics}
\label{subsection:metrics-quality}

Assessing the quality of the KG constructed is important since it is practically impossible to obtain a perfect KG, especially when this latter is very large and populated by automatic approaches from multiple data sources or by manual approaches where human contributors are not necessarily familiar with KGs and have different levels of expertise.
Furthermore, the world is uncertain and knowledge is constantly evolving.
To evaluate a KG, we can rely on five quality dimensions~\cite{WangChen2021, xue2023, hogan2021, Hofer2023}: completeness, accuracy, timeliness, availability, and redundancy.

\noindent\textbf{Completeness} refers to the coverage of knowledge for the specific domain the KG is intended to represent.

\noindent\textbf{Accuracy} corresponds to the correctness of facts in the KG.
In~\cite{Weikum2021}, Weikum \textit{et al.} define some metrics to assess the quality of a KB such as \textit{precision} that captures the accuracy (these terms are sometimes used to describe the same thing), and \textit{recall} that captures the completeness, in the following way:
\begin{equation*}
    precision(S) = \frac{S\cap GT}{S}
    \quad \text{and}\quad
    recall(S) = \frac{S\cap GT}{GT}
\end{equation*}
where $S$ is a set of statements from the KB to be evaluated, and $GT$ is the ground-truth set for the domain of interest.
To deal with uncertain statements that are associated with a confidence score, a threshold is chosen, for which all statements with a score above this threshold are kept.
They also provide an evaluation method that involves uniformly taking a sample of statements and representative of the KB and evaluating it, for example manually, where several annotators may be involved and a consensus or large majority must be found for each annotation.

\noindent\textbf{Timeliness} represents how up-to-date the KG is. The KG can contain temporal facts or facts that evolve and are valid only over a fixed period of time.

\noindent\textbf{Availability} measures the access to KG data, involving its querying and representation.

\noindent\textbf{Redundancy} assesses whether different statements express the same fact, which may entail an entity resolution task.

Another aspect of data quality is the preservation and representation of its provenance and certainty in the form of metadata, which can be used for questions of data selection in relation to both source and quality.
The metadata can also support knowledge fusion approaches by taking them as prior knowledge, as we describe in Section~\ref{section:kg-fusion}.
Other metrics are proposed in the survey~\cite{WangChen2021} for each quality dimension.

\section{Knowledge Graph refinement}
\label{section:kg-refinement}

As presented in Section~\ref{section:kg-construction}, several approaches can be used to extract knowledge from heterogeneous sources (\textit{e.g.,} tables, texts, databases, or human effort) to populate a KG.
The advantage of using multiple sources is two-fold: to ensure knowledge coverage and to identify inconsistencies by leveraging collective wisdom~\cite{dongSaha2012}.
However, the world is uncertain and data sources are of varying quality leading to uncertain knowledge, which we need to handle in the integration process \textit{w.r.t.} quality dimensions listed in Section~\ref{subsection:metrics-quality}.
The causes of uncertainty are presented in Section~\ref{subsubsection:knowledge-deltas}.
Then, we describe our theoretical data integration pipeline for dealing with knowledge uncertainty to enrich a KG in Section~\ref{subsection:requirements}.

\subsection{Knowledge Deltas}
\label{subsubsection:knowledge-deltas}

The uncertainty is everywhere in information and can take the form of knowledge deltas between data sources, according to~\cite{djebri2019}, we adopt this definition of uncertainty in this survey.
We distinguish two types of uncertainty: epistemic, \textit{i.e.,} knowledge about a piece of information is incomplete or unknown; and ontic, \textit{i.e.,} uncertainty is inherent in the information~\cite{walker2003}.
The possible causes of uncertainty are~\cite{anand2022, walker2003}: \textit{(i)} a lack of knowledge; \textit{(ii)} a semantic mismatch or a lack of semantic precision and \textit{(iii)} a lack of machine precision.

Indeed, when building a KG from heterogeneous sources, some kind of deltas of knowledge between sources may appear.
These deltas of knowledge can occur between two data sources on the same subject, \textit{e.g.,} differences in granularity and contradictions.
It is also possible that a data source contradicts itself, a possible way to detect these deltas is to compare the data source to itself by ``reflecting on data patterns or extrapolation to complete missing information and/or detect wrong ones'' according to~\cite{djebri2022}.
On the other hand, duplicates can also occur if the two data sources provide exactly the same knowledge, which needs to be managed for reasons of scalability and KG quality.

\begin{figure*}[h]
	\centering
	\includegraphics[width=\textwidth]{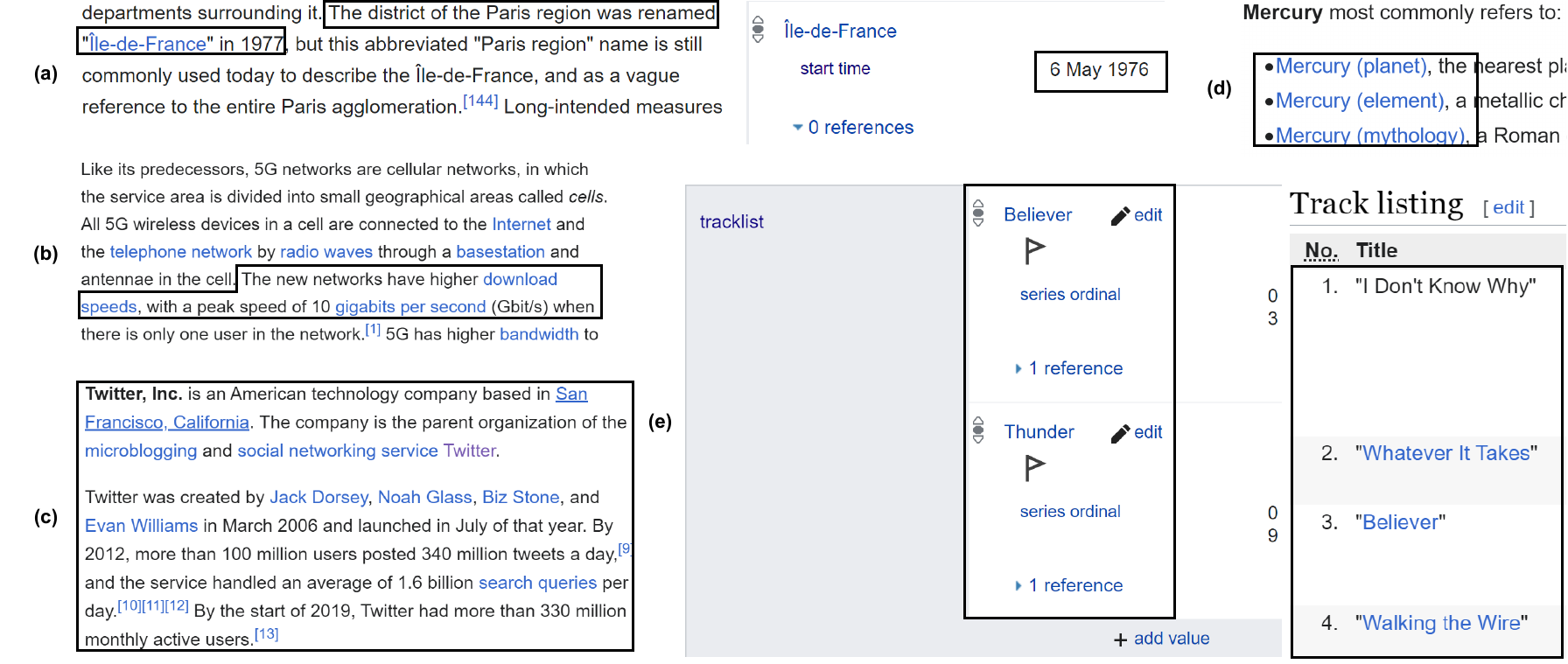}
	\caption{Illustration of the different possible deltas about some topics between English Wikipedia and Wikidata: (a) invalidity + vagueness, (b) fuzziness, (c) timeliness, (d) ambiguity, and (e) incompleteness.}
	\label{fig:delta-examples}
\end{figure*}

We use the same definitions of knowledge deltas as~\cite{djebri2022}, and illustrate them with examples from Wikipedia and Wikidata, depicted in Figure~\ref{fig:delta-examples}.
Suppose that $t$ is a statement. Among the possible knowledge deltas, we find six causes:
\begin{itemize}
    \item \textbf{Invalidity:} $t$ is invalid.
    As illustrated in Figure~\ref{fig:delta-examples} (a), the Wikipedia text of the figure provides incorrect information: the date of renaming of the Paris region to ``île-de-France'' is invalid in the Wikipedia page\footnote{\url{https://en.wikipedia.org/w/index.php?title=Paris&oldid=1197869134}};
    \item \textbf{Vagueness:} $t$ provides vague, imprecise information.
    As depicted in Figure~\ref{fig:delta-examples} (a), the date mentioned on Wikipedia is more vague than the date provided by Wikidata\footnote{\url{https://www.wikidata.org/w/index.php?title=Q90&oldid=2058313448}} for the ``located in the administrative territorial entity'' property which contains additional information such as the day, month, and year;
    \item \textbf{Fuzziness:} $t$ states a fuzzy truth, where the range of values is imprecise itself.
    As depicted in Figure~\ref{fig:delta-examples} (b), the Wikipedia article\footnote{\url{https://en.wikipedia.org/wiki/5G}} about 5G claims that the network has higher download speeds peaking at 10 Gbit/s without specifying a lower bound;
    \item \textbf{Timeliness:} a data source can provide the statement $t$ which is no longer valid at the current time, unlike another source, which can provide an updated version of $t$. 
    As in Figure~\ref{fig:delta-examples} (c), on the Wikipedia page\footnote{\url{https://en.wikipedia.org/w/index.php?title=Twitter,_Inc.&oldid=1087087372}} of May 10, 2022, ``Twitter'' has not yet been renamed to ``X''. This information has now been changed, otherwise there would have been an updating issue;
    \item \textbf{Ambiguity:} $t$ provides multiple interpretations.
    As shown in Figure~\ref{fig:delta-examples} (d), Mercury\footnote{\url{https://en.wikipedia.org/wiki/Mercury}} can be a planet, an element, or a god in mythology;
    \item \textbf{Incompleteness:} $t$ provides incomplete information.
    As in Figure~\ref{fig:delta-examples} (e), the track listing of the album ``Evolve'' by the group Imagine Dragons on Wikidata\footnote{\url{https://www.wikidata.org/w/index.php?title=Q29868187&oldid=2009666363}} contains fewer songs than in Wikipedia\footnote{\url{https://en.wikipedia.org/w/index.php?title=Evolve_(Imagine_Dragons_album)&oldid=1197244329}}.
\end{itemize}

\begin{figure*}[h]
	\centering
	\includegraphics[width=0.8\textwidth]{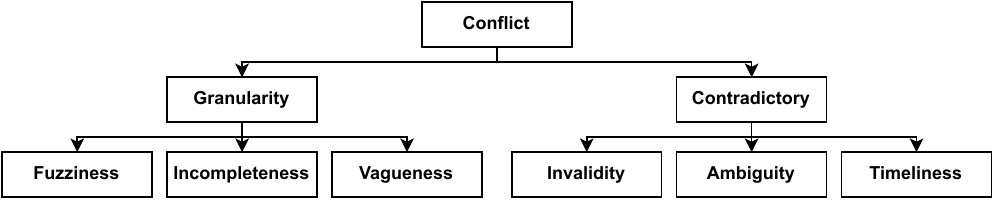}
	\caption{Our proposed classification of knowledge deltas by type of resulting conflict (difference of granularity or contradiction).}
	\label{fig:deltas-classification}
\end{figure*}

The creation of knowledge deltas can be involuntary or voluntary.
An involuntary delta could be the result of uncertain knowledge about a domain (\textit{e.g.,} popular science article \textit{vs} expert article), a typing error, or a data source not up-to-date.
A voluntary delta could simply stem from sabotage by a malicious person (for example, spreading fake news).
Deltas are closely related to the quality dimensions of a KG, since they have a direct impact on them. 
For example, a delta due to the invalidity of an information from a data source directly affects the accuracy of a KG.
We propose to classify these kinds of deltas which lead to conflicts in two classes, namely \textit{Granularity} that stands for a difference between two data sources in the specificity of knowledge and \textit{Contradictory} that stands for an incompatibility of knowledge as depicted in Figure~\ref{fig:deltas-classification}.
We classify \textit{Fuzziness}, \textit{Incompleteness}, and \textit{Vagueness} deltas in the granularity category.
These deltas lead to different levels of specificity between knowledge of two data sources.
This knowledge is not necessarily false, but may be in conflict \textit{e.g.,} a city \textit{vs.} a country to describe the location of an event.
On the other hand, \textit{Invalidity}, \textit{Ambiguity}, and \textit{Timeliness} deltas lead to contradictory knowledge, where some parts of the knowledge are necessarily false.

In~\cite{benslimane2016, zadeh2005}, the authors assume that uncertainty is a common feature of the knowledge we handle daily.
In this sense, exploiting uncertain data sources by ignoring uncertainty to enrich a KG would impact downstream applications of the graph.
The life-cycle for exploiting uncertain data sources requires the measure, the quantification, and the integration of uncertainty in the KG.
In such a view, the uncertainty should be considered anywhere in the pipeline of data integration, including its representation within the KG.
We present our ideal data integration pipeline that tackle the aforementioned requirements in the following section.

\subsection{Requirements for an ideal data integration pipeline}\label{subsection:requirements}
All ways of enriching a KG (\textit{e.g.,} crowdsourcing, extraction from texts or tables, etc.) are error-prone methods since a human cannot be an expert on every domain involving mistakes and extraction algorithms rarely achieve perfect accuracy/precision.
Errors can occur at several stages in the data integration process that encompasses extraction, alignment, or fusion.
Probably one of the most natural ways of capturing and quantifying uncertainty caused by knowledge deltas or the reliability of knowledge integration components is to use confidence scores.
As mentioned in Section~\ref{section:kg-construction}, several extraction approaches provide confidence scores about the triples they extract.
For example, each triple outputted by ReVerb~\cite{fader2011} is associated to a confidence score obtained from a logistic regression. 
Another work~\cite{liGrishman2013} focuses on estimating a confidence score for the slot filling task, which consists in filling predefined attributes for entities in a KB population case.
This confidence score is intended to support the aggregation of values from different slot filling systems. 
The authors have shown that confidence estimation improves performance of the task and that the correctness of values and estimated confidence are strongly correlated.
In~\cite{wick2013}, the authors estimate confidence scores for an entity alignment task that represents the marginal probability that a set of mentions all refer to a same entity.
Therefore, there is a need to consider these confidence scores and represent them as triple metadata along with their provenance information.

\begin{figure*}[h]
	\centering
	\includegraphics[width=\textwidth]{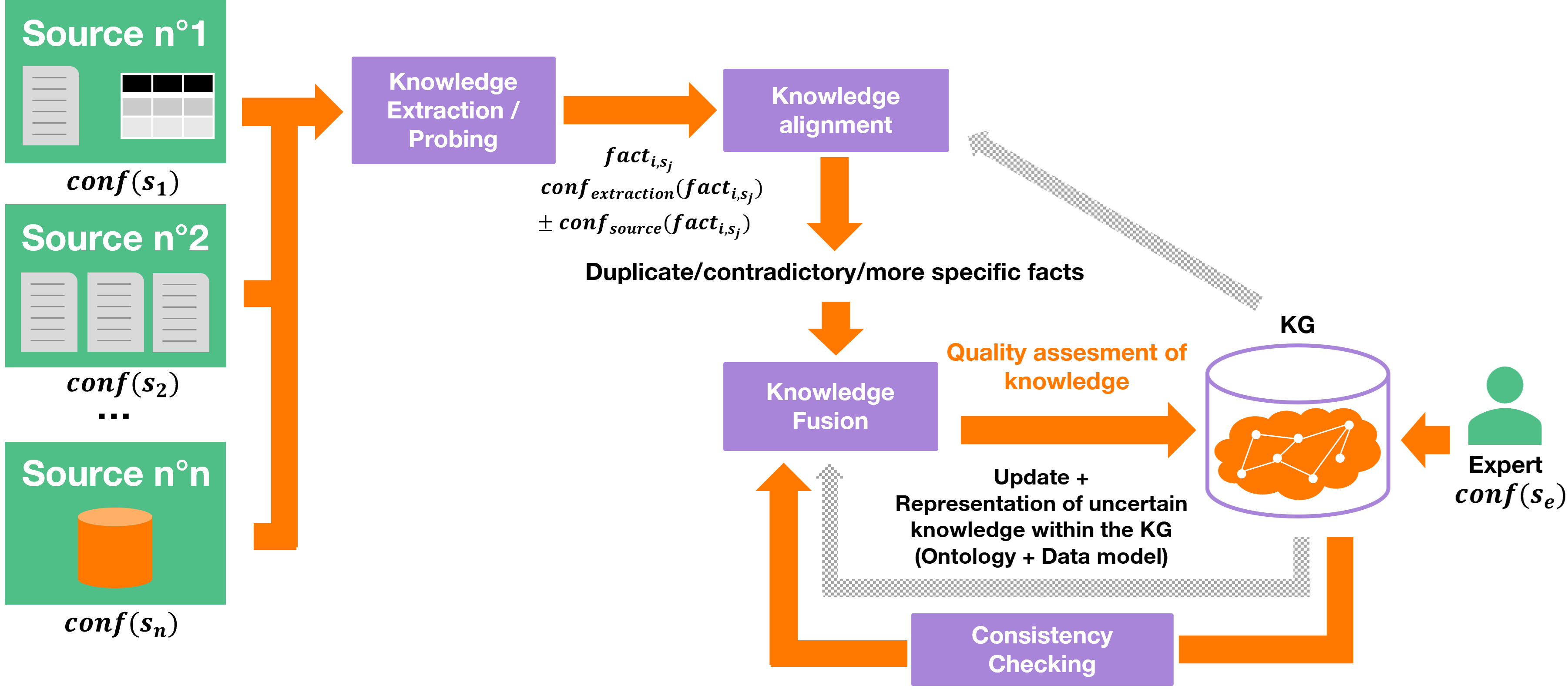}
	\caption{Process of integrating uncertain knowledge extracted from heterogeneous sources, ideally linked to different confidence scores. Three steps form knowledge integration: (1) alignment, (2) fusion, and (3) consistency checking.}
	\label{fig:knowledge-integration}
\end{figure*}

From this perspective, we propose an ideal pipeline of data integration from heterogeneous sources depicted in Figure~\ref{fig:knowledge-integration}.
In the literature, knowledge integration after extraction is often described in two modules~\cite{bleiholder2009}: \textit{Knowledge Alignment} and \textit{Knowledge Fusion}.
In this pipeline, we propose a third module called \textit{Consistency Checking}, which actually takes place after the data integration and identifies and repairs inconsistencies in the KG, improving future knowledge enrichment.
The inputs of the pipeline are multiple heterogeneous sources whose the final purpose is to feed the KG.
From these data sources $s_{j}$, facts $fact_{i,s_{j}}$ are extracted with different confidence scores such as a confidence score in the fact by the extraction algorithm $conf_{extract}(fact_{i,s_{j}})$, a confidence score in the fact by the source $conf_{source}(fact_{i,s_{j}})$ and a confidence score in the source $conf(s_{j})$.
In addition to these multiple data sources, an expert can also populate the KG with a confidence score $conf(s_{e})$.
Before providing these facts to the KG, several tasks are required due to potential knowledge deltas.
The first task \textit{Knowledge Alignment} is the identification of duplicates, differences of granularity and contradictions among extracted facts and the KG.
The next step \textit{Knowledge Fusion} defines a policy to remove conflicting facts and keeps information as consistent, specific, and complete as possible.
Then, knowledge in the KG is updated with their confidence scores and their provenance information since a user might want to query the KG about confidence of triples \textit{w.r.t.} quality dimensions.
A last step verifies the consistency within the KG.
Since this step is performed after the enrichment of the KG and that this survey is focused on the uncertainty management in the construction of KGs, we do not provide further details on it in the following sections.

The aim of this pipeline is to take into account all the confidence scores in the knowledge alignment and fusion modules.
This is not the case in existing work, where only the confidence scores in the data sources are leveraged by the fusion module.
However, methods for KG completion purpose (\textit{i.e.,} predicting new relations using only the KG itself) taking into account uncertainty in an embedding space has recently been studied.

We describe and formalize the ideal data integration policy with an example.
As input, we provide a set of sources $\mathcal{S}$ that contain a set of facts $\mathcal{F}$ and we obtain a $\mathcal{KG}$ as output.
For each fact belonging to the source $f_{\mathcal{S}}$, this fact is aligned with the $\mathcal{KG}$.
If the fact $f_{\mathcal{S}}$ conflicts with a fact already present in the KG $f_{\mathcal{KG}}$, we proceed as follows:
\begin{enumerate}
    \item[(1)] If $f_{\mathcal{S}}$ is more specific than $f_{\mathcal{KG}}$ then we replace $f_{\mathcal{KG}}$ by $f_{\mathcal {S}}$ and we increase the confidence given to the source $\mathcal{S}$.
    For example, ``Joe Biden is president of a North America (NA) country''
    ``Joe Biden is President of the USA'' $\to$ We keep the most specific fact, \textit{i.e.,} ``Joe Biden is president of the USA''. But it would be interesting if we leverage the information ``NA'' by deviating from it $\to$ ``USA is a country in NA'' to complete the graph;
    \item[(2)] Otherwise, if $f_{\mathcal{S}}$ is the duplicate of $f_{\mathcal{KG}}$, we only add the provenance of $f_{\mathcal{S}}$ and we increase the trust placed in the source $\mathcal{S}$;
    \item[(3)] Otherwise, if $f_{\mathcal{S}}$ is contradictory to $f_{\mathcal{KG}}$, we resolve the conflict by finding the true value and we decrease the source that provides an erroneous fact and increase the source that provides a correct one;
    \item[(4)] Else, if the fact does not conflict with a fact of $\mathcal{KG}$, the fact is added to the KG with the associated confidence scores and provenance.
\end{enumerate}
A formalization of the policy is provided in Algorithm~\ref{algo:policy}.

As mentioned in step (4) in the process above, we need to keep a history of knowledge provenance since the provenance information is necessary for KG quality, but could also be used for future conflict resolution or KG updating~\cite{Ilyas2022}.
For this purpose, there is a normative ontology called PROV-O~\cite{lebo2013} that includes provenance information through three components: a set of classes, properties, and restrictions.
It can be used in RDF-based KGs.
The three main classes of the PROV-O ontology are \textit{prov:Entity}, \textit{prov:Activity}, and \textit{prov:Agent} (\textit{prov:Entity} is something that can be changed by an activity, \textit{prov:Activity} is something that acts upon or with entities, and \textit{prov:Agent} can be a human who performs an activity).

In the following sections, we detail three steps of the pipeline namely: Knowledge alignment, Knowledge Fusion, and Uncertainty Representation within the KG.
We propose a section that describes uncertain KG embedding methods for KG completion and confidence prediction tasks (Section~\ref{section:uncertainty-consideration}) before presenting the aforementioned steps.
Knowledge alignment is discussed in Section~\ref{section:kg-alignment}.
In Section~\ref{section:kg-fusion}, we summarize knowledge fusion methods.
Finally, we explore the different mechanisms available for representing triple uncertainty in a KG in Section~\ref{section:uncertainty-representation}.

\begin{algorithm}[h]
	\caption{Data integration policy}\label{algo:policy}
	\begin{algorithmic}
		\REQUIRE A set of facts $\mathcal{F}$ from a source $\mathcal{S}$ with confidence $conf(\mathcal{S})$ where each fact $fact_{\mathcal{S}}$ is associated to a confidence by the algorithm of extraction $conf_{extract}(fact_{\mathcal{S}})$ and a confidence by the source $conf_{source}(fact_{\mathcal{S}})$, $conf(fact_{\mathcal{S}}) = (conf(\mathcal{S}), conf_{extract}(fact_{\mathcal{S}}), conf_{source}(fact_{\mathcal{S}}))$, and a $\mathcal{KG}$
		\ENSURE $\mathcal{KG}$ updated with consistent facts of $\mathcal{S}$
		\FOR{$f_{\mathcal{S}} \in \mathcal{F}$}
		\STATE $f_{\mathcal{KG}} \gets \mathrm{align}(\mathcal{KG}, f_{\mathcal{S}}, conf(fact_{\mathcal{S}}))$ 
		\IF{$f_{\mathcal{KG}} \neq \emptyset$}
		\IF{$\mathrm{moreSpecific}(f_{\mathcal{S}}, f_{\mathcal{KG}})$}
		\STATE $\mathrm{replace}(f_{\mathcal{KG}}, f_{\mathcal{S}})$
		\STATE $\mathrm{increase}(conf(\mathcal{S}))$
		\ELSIF{$\mathrm{similar}(f_{\mathcal{S}}, f_{\mathcal{KG}})$}
		\STATE $\mathrm{addSource}(\mathcal{KG}, f_{\mathcal{KG}}, \mathcal{S})$
		\STATE $\mathrm{increase}(conf(\mathcal{S}))$
		\ELSIF{$\mathrm{contradictory}(f_{\mathcal{S}}, f_{\mathcal{KG}})$}
		\STATE $\mathrm{decrease}(conf(\mathcal{S}))$
		\ENDIF
		\ELSE
		\STATE $\mathcal{KG} \gets \mathcal{KG} \cup \{f_{\mathcal{S}}\}$
		\ENDIF
		\ENDFOR
	\end{algorithmic}
\end{algorithm}

\section{Uncertain Knowledge Graph Embedding}
\label{section:uncertainty-consideration}

Embedding methods enable KG representation in a $n$-dimensional vectorial space, \textit{i.e.,} its entities and relations are $n$-vectors.
These embeddings attempt to preserve the structural properties of the graph and thus make it easy to manipulate the graph for machine learning applications such as link prediction, completion, or node classification~\cite{ji2021}.
A wide range of embedding models have emerged such as TransE~\cite{bordes2013}, DistMult~\cite{yang2015}, ComplEx~\cite{trouillon2016}, RotatE~\cite{sunDeng2019}, neural networks applied to graphs such as RGCN~\cite{schlichtkrull2018}, or GCN~\cite{kipf2017}.
In addition, embeddings are also increasingly used in the construction of KGs (for example, for knowledge alignment~\cite{fanourakis2023}, or other tasks for KG refinement~\cite{hogan2022}).
Most embedding approaches do not include the uncertainty of knowledge in their models.
However, when constructing KBs, the knowledge is often uncertain or noisy and not taking into account uncertainty during the representation learning can imply a bias in its representation and impact further applications.
Given the importance of embedding methods in both KG applications and construction, we consider that it is useful to gather such methods that include uncertainty expressed in terms of a confidence score in their modeling.
This section describes some of these models and the datasets used to evaluate them.

\subsection{Uncertain KG embedding models}
In this paper, we formalize uncertain KGs as follows. An {\it uncertain KG} (UKG) is represented as a set of weighted triples $\mathcal{G} = {(s, p, o, s_{t})}$, where $(s, p, o)$ is a triple representing a fact and $s_{t} \in [0, 1]$ is a confidence score for this fact to be true.
The uncertainty linked to the triples in the KG relies on the plausibility of the triples, but most KG embedding (KGE) methods do not consider this information in their modeling, making the assumption that all triples are deterministic.
Such an assumption does not reflect the reality where many triples are uncertain due to the reasons described in Section~\ref{section:kg-refinement}.
Table~\ref{tab:ukge-recap} summarizes the UKG embedding approaches with their associated tasks, scoring function, the year of publication, and the datasets on which the experiments were conducted.
We can notice that uncertain graph embeddings have only been recently studied.

\textbf{UKGE~\cite{chen2019}} improves traditional KGE models by using the Probabilistic Soft Logic (PSL) framework to infer confidence scores for unseen relational triples.
Thus, UKGE encodes the KG according to confidence scores for observed and unseen triples.
They map the scoring function results into confidence scores using two different mapping functions, namely a logistic function or the bounded rectifier function.
For the relation fact classification, global ranking, and confidence prediction tasks, UKGE outperforms the deterministic KG embedding models such as TransE, DistMult, ComplEx and the URGE model on CN15k, NL27k and PPI5k datasets.

\textbf{SUKE~\cite{wang2021}} argues that UKGE does not make full use of the structure information of the fact.
Therefore, to improve this, SUKE has two components: an evaluator and a confidence generator.
The evaluator defines a structure score and an uncertainty score for each fact to capture the rationality of triples.
The generator outputs confidence scores for triples from the uncertainty score computed by the evaluator.
The plausibility of facts are computed with DistMult~\cite{yang2015} scoring function, then it applies a different mapping function with two parameters for the structural score and the uncertain score before merging them.
The confidence generator only uses the uncertainty score to approximate the true confidence value of triples.

\textbf{BEURRE~\cite{chen2021}} models entities as probabilistic boxes and relations between two entities as an affine transformation.
The confidence score of the relation between two entities is represented as the volume of the intersection of their boxes.
Constraints such as transitivity and composition are inserted into the modeling of embeddings to preserve these properties on relations in the embedding space.
These constraints act as a loss regularization in the global loss function.
Then, embeddings are trained by optimizing a loss function for a regression task and a regularization loss to apply transitivity and composition constraints.

\textbf{GTransE~\cite{kertkeidkachorn2019}} embeds uncertainty through a translational model by expanding the well known TransE~\cite{bordes2013}.
The uncertainty is included at the level of the loss function on a hyperparameter of the margin loss function when training the embeddings:
\begin{align*}
    \mathcal{L} &= \sum_{(h,r,t,s)\in Q}^{}\sum_{(h',r,t',s)\in Q'}^{}\left[ f(h,r,t)-f(h',r,t')+s^{\alpha}M \right]_{+}\\
    &= 
    \sum_{(h,r,t,s)\in Q}^{}\sum_{(h',r,t',s)\in Q'}^{}\max{0, f(h,r,t)-f(h',r,t')+s^{\alpha}M }
\end{align*}
where $(h,r,t,s)$=(head, relation, tail, confidence score), M a margin parameter and $\left[x \,\right]_{+}$ is the positive part of $x$.
The scoring function $f$ corresponds to L1 or L2-norm.
Thus, with this loss function if a triple $(h,r,t)$ has a high confidence score it will tend to respect $h + r = t$ otherwise the entity $t$ will tend to move away from $h + r$.
Before GTransE, the same authors introduced \textbf{CTransE~\cite{kertkeidkachorn2019ctranse}} that is closely the same model but without the hyperparameter $\alpha$ in power of the confidence score.

\textbf{IIKE~\cite{fan2016}} models the confidence in the embedding space through a probabilistic model.
The authors propose an embedding model that takes uncertainty into account by minimizing a loss function to fit the output confidence of triples acquisition (\textit{e.g.,} NELL, or crowdsourcing) to the scoring function of the triples given by a probability function.
The plausibility of a triple is modeled as a joint probability of the head of the entity, the relation and the tail $Pr(h,r,t)$ depending on $Pr(h|r,t)$, $Pr(r|h,t)$, and $Pr(t|h,r)$.
For the loss function, the authors minimize the difference between the logarithm of the triple probabilities and the logarithm of the confidence of the knowledge extraction. Then apply stochastic gradient descent to refine embeddings at each iteration.

\textbf{PASSLEAF~\cite{chenYeh2021}} decomposes the model in two parts: a confidence score prediction framework that adapts the score function among existing ones, \textit{e.g.,} ComplEx~\cite{trouillon2016} or RotatE~\cite{sunDeng2019} and a semi-supervised learning framework.

For the UKG completion task, each relation must have enough training examples to perform correctly. \textbf{GMUC~\cite{zhangWu2021}} addresses the few-shot UKG completion task for long-tail relations.
GMUC learns a Gaussian similarity metric that allows missing facts and their confidence scores to be predicted from a few training examples.
The model encodes a support set containing a few facts with their confidence scores and a query into multidimensional Gaussian distributions.
The query consists of pairs \textit{(head, relation)} where tail and score must be predicted.
Then a Gaussian matching function is used to generate a similarity distribution between the query and the support set.
GMUC outperforms UKGE model on link prediction and confidence prediction on NL27K and three NL27K-derived datasets with added noise.

\textbf{UOKGE~\cite{boutouhamiZhang2019}} learns embeddings of uncertain ontology-aware KGs according to confidence scores.
It encodes an instance $e_{i} \in E$ as a point represented by a n-dimensional vector, a class $c_{i} \in C$ as a sphere $s_{i}(c_{i}, \rho_{i})$ where $c_{i}$ is the center of the sphere and $p_{i}$ is the radius, and a property $p_{i} \in P$ as a sphere $s_{i}((p_{i}^{d},p_{i}^{r}),\varrho_{i})$ where $(p_{i}^{d},p_{i}^{r})$ is the center of the sphere with $p_{i}^{d}$ representing the domain, $p_{i}^{r}$ representing the range, and $\varrho_{i}$ is the radius.
Then, it introduces the following mapping function that allows changing the scale of values between 0 and 1 to represent the uncertainty.
Six distinct gap functions for six types of relation are defined to encode uncertainty for: type, domain, range, subclass, sup-property and remaining properties.
Then, it minimizes the mean squared error between the confidence score and those gap functions.

\textbf{FocusE~\cite{paiCostabello2021}} improves KG embeddings with numerical values on edges by taking action between the scoring function of traditional models (\textit{e.g.,} TransE, ComplEx, or DistMult) and the loss function.
They introduce numerical values on edges in a way that maximizes the margin between the scores associated with true triples and their corruptions.
Given the score function of an embedding model $f(t)$ with $t$ a triple, they use a nonlinear softplus $\sigma$ in order that the score provided by $f(t)$ is greater or equal to zero:
\begin{equation*}
    g(t)=\sigma(f(t)) = ln(1+e^{f(t)})\ge 0
\end{equation*}
Then, the numerical value associated to an edge is expressed through $\alpha$ in the following way: 
\begin{equation*}
    \alpha = 
    \begin{cases}
       \beta+(1-w)(1-\beta) \text{ if $t$} \\
       \beta+w(1-\beta) \text{ if $t^{-}$}
    \end{cases}
\end{equation*}
where $\beta$ is a hyperparameter acting on the importance of the topological structure of the graph and $w$ is the numerical value on the edge. 
Then, the final function of FocusE is the following: $h(t) = \alpha g(t)$.

\textbf{ConfE~\cite{zhaoHou2021}} encodes the tuples $(e, \tau)$ with $e$ an entity and $\tau$ an entity type by taking into account the uncertainty in each tuple. 
They consider entities and entity types as two different things in a KG and learn embeddings of entities and entity types in two distinct spaces with an asymmetric matrix to model their interactions.
The scoring function is defined as $G(e,\tau)=e^{\top}M\tau$ where $M$ is the asymmetric matrix.
And uncertainty is included within the loss function: 
\begin{equation*}
    \mathscr{L}=\sum_{(e,\tau)\in \mathscr{H}}^{}\sum_{(e',\tau')\in \mathscr{H'}}^{}\text{max}(0,\gamma-G(e,\tau)+G(e',\tau')).C(e,\tau)
\end{equation*}
where $\mathscr{H}$ is the set of entities and their type and $\mathscr{H}'$ the set of corrupted tuples.

\textbf{CKRL~\cite{xieLiu2018}} introduces multiple levels of confidence, namely a local triple confidence, a global path confidence, a prior path confidence, and an adaptive path confidence.
These confidence scores are integrated into the energy function following designed: 
\begin{equation*}
    E(T) = \sum_{(h,r,t)\in T}^{}E(h,r,t)\cdot C(h,r,t)
\end{equation*}
where $E(h,r,t)=\left\| h + r - t \right\|$ and $C(h,r,t)$ the triple confidence score aggregating all levels of confidence.

\textbf{WaExt~\cite{Kong}} embeds triples of a KG by incorporating the weight $w$ associated to an edge $(h,r,t)$ in the scoring function in the following manner: 
\begin{equation*}
    f_{w}(h,r,t,w) = g(w) \cdot f(h,r,t)
\end{equation*}
and then minimizes a margin ranking loss function.

\textbf{Wang \textit{et al.}~\cite{WangWu2022}} model each entity and each relation as a multidimensional Gaussian distribution $\mathcal{N}(\mu, \Sigma)$ where $\mu$ is a mean vector representing its position and $\Sigma$ is a diagonal covariance matrix representing its uncertainty.

\textbf{MUKGE}~\cite{liuZhang2024} aims to improve the generation of unseen facts for KGE training.
The authors argue that PSL cannot take advantage of global multi-path information involving information loss to estimate the confidence of unseen facts.
Indeed, PSL only considers information from simple logical rules with a path length of two as used in UKGE~\cite{chen2019} (\textit{e.g.,} $(college, synonym, university)$ $\wedge$  \textit{(university, synonym, institute)} $\to$ \textit{(college, synonym, institute)}) and other paths in the graph between the subject and object of the inferred relation are not taken into account.
To solve this issue, MUKGE introduces an algorithm called Uncertain ResourceRank to infer confidence scores for unseen triples based on the relevance of the entity pairs (subject, object).
The relevance of an entity pair is computed \textit{w.r.t.} the directed paths between subject and object in the KG.
MUKGE uses circular correlation as scoring function, and either applies the sigmoid or the bounded rectifier function as the function to obtain the triple confidence. 
Then the authors design the loss function to fit each positive triple to its corresponding confidence score.
The authors assess their model on confidence prediction, relation fact ranking, and relation fact classification. 
For these three tasks, MUKGE outperforms the BEURRE and UKGE models, with a focus on asymmetrical relations. 
However, for all relation types, performance is competitive with those of other models, except for confidence prediction, where MUKGE is the best alternative.

\begin{table*}
	\begin{adjustbox}{width={\textwidth},totalheight={0.75\textheight},keepaspectratio}
		\begin{tabular}{lcccc}
			\hline
			Model & Scoring function & Evaluation task & Dataset & Year \\
			\hline
               \multirow{2}{*}{IIKE~\cite{fan2016}} & \multirow{2}{*}{TransE} & Link prediction & NELL & 2016 \\
			&  & Triple classification & FB15K (synthetic) &  \\
			\hline
            \multirow{3}{*}{URGE\cite{Hu2017}} &  & Node Clustering & PPI &  \\
			& Matrix Factorization (proximity preservation) & Node Classification & DBLP & 2017 \\
			&  & k-NN search &  & \\
			\hline
            \multirow{3}{*}{CKRL~\cite{xieLiu2018}} &  & KG Noise Detection &  &  \\
			& TransE & KG Completion & FB15K (synthetic) & 2018 \\
			&  & Triple Classification &  & \\
			\hline
            \multirow{2}{*}{GTransE~\cite{kertkeidkachorn2019}} & \multirow{2}{*}{TransE} & KG Completion & FB15K-237 (synthetic) & 2019 \\
			&  &  & NELL (synthetic) & \\
			\hline
			\multirow{3}{*}{UKGE~\cite{chen2019}} &  & Confidence prediction & CN15K & \\
			& DistMult & Relation fact ranking & NL27K & 2019 \\
			&  & Relation fact classification & PPI5K &  \\
			\hline
            \multirow{2}{*}{UOKGE~\cite{boutouhamiZhang2019}} & \multirow{2}{*}{TransE} & Confidence Prediction & CN15K & 2019 \\
			& & Triples Classification &  & \\
			\hline
            \multirow{3}{*}{SUKE~\cite{wang2021}} &  & Link prediction & CN15K &  \\
			& DistMult & Fact classification & NL27K & 2021 \\
			& & & PPI5K  &  \\
			\hline
			\multirow{2}{*}{BEURRE~\cite{chen2021}} & \multirow{2}{*}{Gumbel boxes} & Confidence prediction & CN15K & 2021 \\
			&  & Relation fact ranking & NELL27k &  \\
			\hline
			\multirow{5}{*}{PASSLEAF~\cite{chenYeh2021}} &  &  & PPI5K & \\
			& RotatE & Tail Entity Prediction & NL27K & \\
			& ComplEx & Confidence Prediction & CN15K & 2021 \\
			& DistMult &  & WN18RR & \\
			&  &  & FB15K237 & \\
			\hline
			\multirow{2}{*}{GMUC~\cite{zhangWu2021}} & \multirow{2}{*}{Minimum Similarity} & Link Prediction & NL27K & 2021 \\
			&  & Confidence Prediction & & \\
			\hline
			\multirow{2}{*}{ConfE~\cite{zhaoHou2021}} & \multirow{2}{*}{RESCAL} & Entity Type Noise Detection & FB15kET (synthetic) & 2021 \\
			&  & Entity Type Prediction & YAGO43k (synthetic) & \\
			\hline
			\multirow{4}{*}{FocusE~\cite{paiCostabello2021}} & \multirow{4}{*}{Modifiable Scoring Layer} & \multirow{4}{*}{Link Prediction (High-Valued Links)} & O*NET20K & 2021 \\
			&  &  & CN15K & \\
			&  &  & NL27K & \\
			&  &  & PPI5K & \\
			\hline
			\multirow{3}{*}{WaExt~\cite{Kong}} & TransE &  & CN15K & 2022 \\
			& TransH & Link Prediction & NL27K &  \\
			& DistMult & Triple Classification & PPI5K &  \\
			& ComplEx &  &  &  \\
			\hline
			\multirow{2}{*}{Wang \textit{et al.}~\cite{WangWu2022}} & \multirow{2}{*}{Similarity} & Confidence Prediction & CN15K & 2022 \\
			&  & Tail Entity Prediction & NL27K &  \\
			\hline
           \multirow{2}{*}{MUKGE~\cite{liuZhang2024}} & \multirow{2}{*}{Circular correlation} & Confidence Prediction & CN15K & 2024 \\
			&  & Relation Fact Ranking & NL27K &  \\
            &  & Relation Fact Classification & PPI5K &  \\
			\hline
		\end{tabular}
	\end{adjustbox}
	\caption{Uncertain embedding models taking into account at least one numeric value on edges with their scoring function, tasks handled, datasets on which they are evaluated, and the year of the publication.}
	\label{tab:ukge-recap}
\end{table*}

\subsection{Datasets with numerical values on edges}
\label{subsubsection:uncertain-datasets}

In the literature, five datasets that come from uncertain KBs are commonly used in UKG completion or confidence prediction tasks presented in the previous section~\cite{paiCostabello2021}.
\textbf{CN15K} is a subset of ConceptNet (presented in Section~\ref{subsubsection:open-kgs}) where the numerical values correspond to the uncertainty of the triples~\cite{Speer2016}. 
The confidence scores for each triple are computed \textit{w.r.t.} the number of sources and their reliability.
\textbf{NL27K} is a subset of NELL dataset (presented in Section~\ref{subsubsection:web}) where the confidence scores are computed and refined by an Expectation Maximization (EM) algorithm and a semi-supervised learning method.
\textbf{PPI5K} is a KG that represents the protein-protein interactions, and the numerical values correspond to the confidence relation~\cite{szklarczyk2017}.
\textbf{O*NET20K} is a dataset introduced by~\cite{paiCostabello2021} which includes descriptions about jobs and skills. 
The numerical values represent the strength of the relations.
Some embedding models also generate their own synthetic noisy datasets with fictitious confidence scores following probability distributions.

\section{Knowledge Alignment}
\label{section:kg-alignment}

Knowledge alignment, \textit{a.k.a.} knowledge resolution or knowledge matching, is the process of finding relationships or correspondences between entities of different ontologies~\cite{euzenat2013}.
It is the first step in the pipeline after knowledge acquisition, which identifies candidate entities for knowledge fusion.
For example, in Figure~\ref{fig:knowledge-fusion-galaxy} the entity ``Galaxy S23'' of both graphs refers to the same entity in real world but are stemmed from two different sources.
This task copes with the ``redundancy'' quality dimension {(Section~\ref{subsection:metrics-quality}).
Whether at instance level or at ontology level, many works tackle the knowledge alignment task.
This section aims to provide an overview of the knowledge alignment task and approaches by gathering the various existing surveys~\cite{fanourakis2023, Euzenat2007, sun2020}.

The authors of \cite{euzenat2013} distinguish different types of matching that include semantic or syntactic approaches, \textit{e.g.,} string-based, language-based, subgraph-based, rule-based, embedding-based, or relational-based approaches.
An example of such a rule-based method is the following, by ~\cite{jiang2016}. 
For each relation $R_{k}$ over the two domains,  define 
\begin{align}
	R_{k}(a,b)\wedge \neg R_{k}(a',b')\Rightarrow a\not\equiv a' \vee b \not\equiv b'\\
	R_{k}(a,b)\wedge R_{k}(a',b')\Rightarrow a\equiv a'\wedge b\equiv b'\\
	\forall a,b \in o, \, a',b' \in o',
\end{align}
where $R_{k}$ is a relation present in both graphs.
To reduce complexity and avoid scalability issues, some approaches use blocking methods that avoid unnecessary comparisons by gathering entities.
For example, Nguyen \textit{et al.}~\cite{nguyen2020}  propose different strategies of blocking based on entities' description such as \textit{token blocking}, \textit{i.e.,} entities in the same cluster share at least one common token in their description, \textit{attribute clustering blocking}, \textit{i.e.,} clusters the entities in the same group if their attributes are similar, and \textit{prefix-infix(-suffix) blocking}, \textit{i.e.,} exploits the pattern in the description of the URI (\textit{e.g.,} URI infix) to create new blocks.
After an optional blocking step, knowledge alignment methods are performed.
Among them, \cite{fanourakis2023} distinguish three methods: Sharing, Swapping, and Mapping.
\textit{Sharing} updates the entity embeddings produced by the embedding module according to the available similarity evidence of entities.
\textit{Swapping} updates the entity embeddings produced by the embedding module according to the available similarity evidence of entities but adds positive triples by leveraging aligned pairs \textit{e.g.,} $(h, h’)$, $(t, t’)$ $\to$ $(h’, r, t) + (h, r, t’)$.
\textit{Mapping} learns a linear transformation between the two embedding spaces of aligned KGs.

Furthermore, alignment approaches are diverse and varied, some of them leverage attributes of entities, or use only relations between entities where different depths of context (\textit{e.g.,} neighboring entities) are considered, while for other the path in the graph is an important aspect.
We provide Table~\ref{tab:alignment-recap} that summarizes these different existing alignment approaches to get an overview strongly inspired by~\cite{fanourakis2023, sun2020}.

\begin{table}
    \begin{adjustbox}{width={\textwidth},totalheight={0.65\textheight},keepaspectratio}%
    \begin{tabular}{lccccc}
         \toprule
         \multicolumn{1}{l}{\multirow{2}{*}{Model}} & \multicolumn{3}{c}{Embedding} & \multicolumn{1}{c}{\multirow{2}{*}{Method}} & \multicolumn{1}{c}{\multirow{2}{*}{Learning}} \\
         \cmidrule{2-4}
         \multicolumn{1}{c}{} & \multicolumn{1}{c}{One-hop} & \multicolumn{1}{c}{Multi-hop} & \multicolumn{1}{c}{Path} & \multicolumn{1}{c}{} & \multicolumn{1}{c}{} \\
         \midrule
         MTransE~\cite{MTransE} & $\bullet$ & & & Mapping & Supervised \\
         IPTransE~\cite{IPTransE} & & & $\bullet$ & Sharing & Semi-supervised \\
         JAPE~\cite{JAPE} & $\bullet$ & & & Sharing & Supervised \\
         BootEA~\cite{BootEA} & $\bullet$ & & & Swapping & Semi-supervised \\
         KDCoE~\cite{kdcoe} & $\bullet$ & & & Mapping & Semi-supervised \\
         NTAM~\cite{ntam} & $\bullet$ & & & Swapping & Supervised \\
         GCNAlign~\cite{gcnalign} & & $\bullet$ & & Mapping & Supervised \\
         AttrE~\cite{attre} & $\bullet$ & & & Sharing & Supervised \\
         IMUSE~\cite{imuse} & $\bullet$ & & & Sharing & Supervised \\
         SEA~\cite{sea} & $\bullet$ & & & Mapping & Supervised \\
         RSN4EA~\cite{rsn4ea} & & & $\bullet$ & Sharing & Supervised \\
         GMNN~\cite{gmnn} & & $\bullet$ & & Swapping & Supervised \\
         MuGNN~\cite{mugnn} & & $\bullet$ & & Mapping & Supervised \\
         OTEA~\cite{otea} & $\bullet$ & & & Mapping & Supervised \\
         NAEA~\cite{naea} & & $\bullet$ & & Swapping & Supervised \\
         AVR-GCN~\cite{avr-gcn} & & $\bullet$ & & Swapping & Supervised \\
         MultiKE~\cite{multike} & $\bullet$ & & & Swapping & Supervised \\
         RDGCN~\cite{rdgcn} & & $\bullet$ & & Mapping & Supervised \\
         KECG~\cite{kecg} & & $\bullet$ & & Mapping & Supervised \\
         HGCN~\cite{hgcn} & & $\bullet$ & & Mapping & Supervised \\
         MMEA~\cite{mmea} & $\bullet$ & & & Sharing & Supervised \\
         HMAN~\cite{hman} & & $\bullet$ & & Mapping & Supervised \\
         AKE~\cite{ake} & $\bullet$ & & & Mapping & Supervised \\
         RREA~\cite{rrea} & & $\bullet$ & & Sharing & Supervised \\
         BERT\_INT~\cite{bert_int} & & $\bullet$ & & Sharing & Supervised \\
         MTransE+RotatE~\cite{mtranse+rotate} & $\bullet$ & & & Sharing & Supervised \\
         \bottomrule
    \end{tabular}
    \end{adjustbox}
    \caption{Recent embedding-based approaches that tackle knowledge alignment task.}
    \label{tab:alignment-recap}
\end{table}

The authors of \cite{fanourakis2023} highlight that BERT\_INT outperforms all models in terms of effectiveness and efficiency overall, especially when the KGs contain highly similar factual information.
In fact, the alignment models that use language models such as BERT\_INT are the most efficient for this task.
The authors of~\cite{fanourakis2023} indicate the critical factors that affect the effectiveness of relation-based and attribute-based alignment methods, for instance:
\begin{itemize}
	\item the depth of neighbors considered;
	\item negative sampling (for training), as the number of negatives are considered, the performances decrease;
	\item depending on inputs KGs to align, \textit{e.g.,} for OpenEA datasets, is not necessary to use attributes information, factual information is sufficient.
\end{itemize}

\section{Uncertain Knowledge Fusion}
\label{section:kg-fusion}

In the previous section, we introduced the knowledge alignment task with a summary of the embedding-based approaches that tackle it.
This task identifies equivalent entities and groups them into different clusters.
The next step is the fusion of the attributes of the entities into the same clusters (as illustrated in Figure~\ref{fig:knowledge-integration}) since they may be redundant, inconsistent, contradictory, or of different granularity.
We first define the task in Section~\ref{subsubsection:fusion-definition}, and we present the different approaches of fusion in Section~\ref{subsubsection:fusion-approaches}.

\subsection{Task definition}
\label{subsubsection:fusion-definition}

The knowledge fusion step consists of studying how to combine various information about the same entity or concept from multiple data sources into a consistent and a unified one regarding the different deltas listed in the Section~\ref{subsubsection:knowledge-deltas}~\cite{Hofer2023, Nikolov2007}.
The authors of \cite{Dong2009} identify three broad goals to be achieved for this challenging task:
\begin{itemize}
	\item Completeness: measures the expected amount of data (number of tuples and number of attributes) at the output of the fusion task;
	\item Conciseness: measures the uniqueness of object representations in the integrated data (number of unique objects and number of unique attributes of objects);
	\item Correctness: measures the correctness of data, \textit{i.e.,} its conformity to the real world.
\end{itemize}

Therefore, data fusion corresponds to resolve conflicts from data with respect to these three goals.
In~\cite{bleiholder2006}, the authors distinguish two types of data conflicts from a data fusion perspective: \textit{contradictions} and \textit{uncertainties}.
He defines \textit{contradictions} as follows: ``a contradiction is a conflict between two or more different non-null values that are all used to describe the same property of an object'' and \textit{uncertainties} as follows: ``an uncertainty is a conflict between a non-null value and one or more null values that are all used to describe the same property of an object''.
We adopt the same definition of \textit{contradictions}, but we do not consider the same definition of \textit{uncertainty}.
We define the second type of conflict as a difference in granularity of knowledge, as illustrated in Figure~\ref{fig:deltas-classification}.

In this survey, ``uncertainty'' is a more generic term whose sources lie in knowledge deltas and the inaccuracy of each step in the knowledge integration pipeline, including knowledge acquisition.
Indeed, when we integrate knowledge from several sources, the quality of the information varies, and we need to determine the trustworthy information by performing a Truth Inferring (TI) task.

According to Rekatsinas~\cite{Rekatsinas2017} different TI strategies can be adopted.
There are simple strategies that estimate the true values of the entities compared to the other values provided by the sources by applying a majority vote or an average on them.
Then, there are strategies that use the trustworthiness of the sources to quantify the true values of the objects, and it is even possible to establish a precision for each class of object and for each source.
The problem with simple strategies is that they do not take into account the varied quality of data sources~\cite{LiLi2014}, but these are often used to initialize true values to start iterative TI methods.

For example, we consider that we have previously extracted knowledge in triple form \textit{(subject, predicate, object)} about the mobile phone ``Galaxy S23'' from several sources $\mathcal{S}_{1}$, $\mathcal{S}_2$, ..., $\mathcal{S}_{n}$ resulting in the table at the bottom of Figure~\ref{fig:knowledge-fusion-galaxy} where entities have already been aligned.
We also represent the table as a graph for sources $\mathcal{S}_{1}$ and $\mathcal{S}_{2}$.
\begin{figure*}[h]
	\centering
	\includegraphics[width=\textwidth]{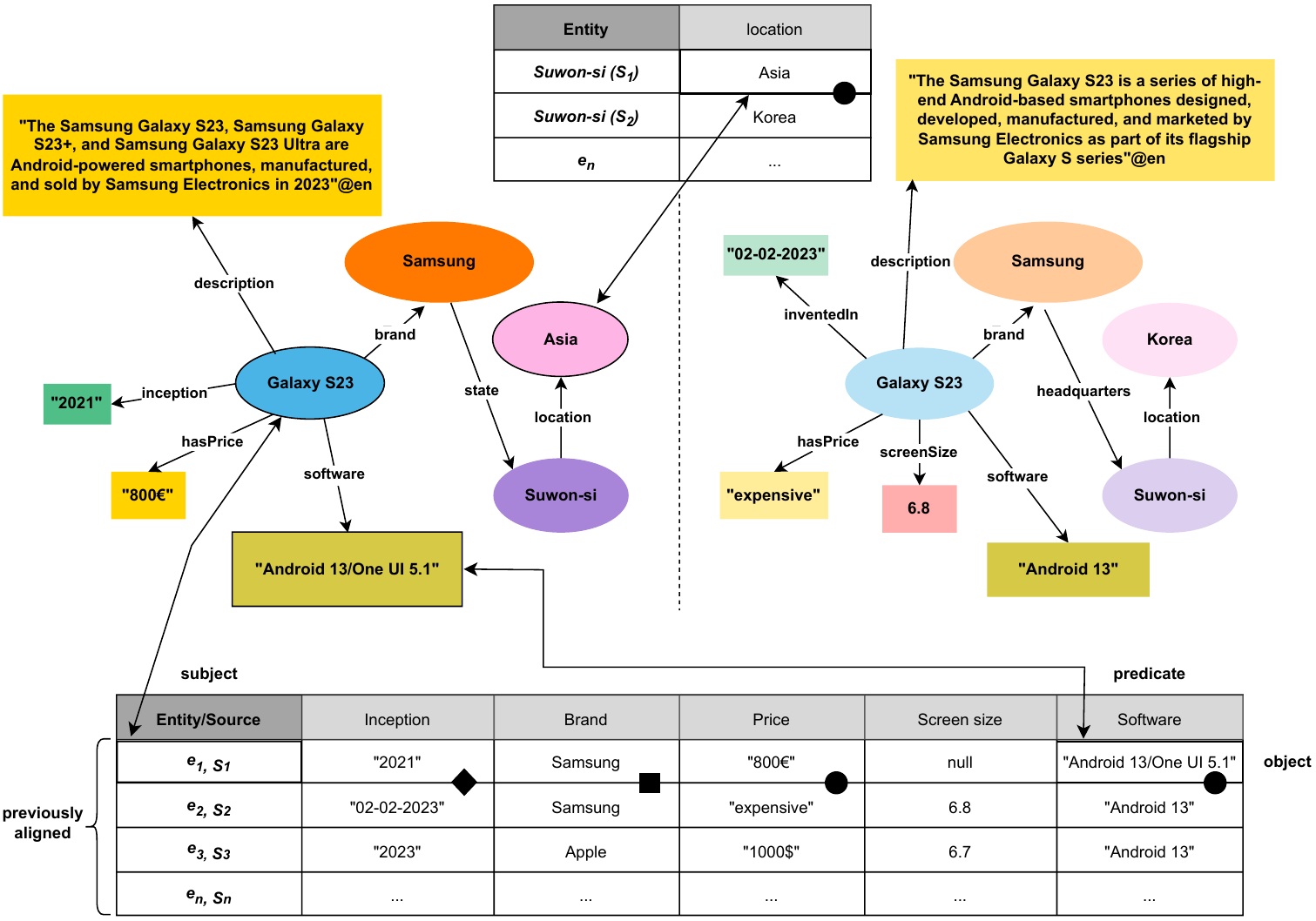}
	\caption[]{Extracted triples represented as graphs from two sources $\mathcal{S}_{1}$ and $\mathcal{S}_{2}$. $\mathcal{S}_{1}$ on the left and $\mathcal{S}_{2}$ on the right. With different types of conflicts:
 contradictory information
 (\raisebox{-1pt}{\begin{tikzpicture}[scale=0.5]
        \draw[color=losange, fill=losange] (0,0.25) -- (0.25,0) -- (0.5,0.25) -- (0.25,0.5) -- cycle;
    \end{tikzpicture}}),
    duplicate
    (\raisebox{-1pt}{\begin{tikzpicture}[scale=0.5]
        \draw[color=carre, fill=carre] (0,0) -- (0.425,0) -- (0.425,0.425) -- (0,0.425) -- cycle;
    \end{tikzpicture}}), and difference of granularity
    (\raisebox{-1pt}{\begin{tikzpicture}[scale=0.5]
        \draw[color=rond, fill=rond] (0,0) circle (7pt);)
    \end{tikzpicture}}).}
	\label{fig:knowledge-fusion-galaxy}
\end{figure*}
Several papers on data fusion use the term ``data item'', \textit{i.e.,} \textit{(entity, attribute, value)} instead of the term ``triple'' to refer to an element to be merged.
However, in practice a data item is equivalent to a triple \textit{(subject, predicate, object)}.
Each row of the table corresponds to an entity of a graph and its attributes, for example the entity $e_{1}$ corresponds to the node ``Galaxy S23'' of the graph and the values associated with $e_{1}$ in the table are the objects of the triples.
These objects are linked to the subject ``Galaxy S23'' by the predicates identified by column headers, as depicted in Figure~\ref{fig:knowledge-fusion-galaxy}.
The data extracted from both sources is almost the same except for the relations $software$ and $inception$ where differences of granularity appear (\textit{e.g.,} the price of the mobile phone).
Source $\mathcal{S}_{3}$ states that the brand of the mobile phone is ``Apple'', contradicting the first two data sources.
Another example of contradiction is the invalidity of the phone's inception date provided by $\mathcal{S}_{1}$.
We can also distinguish two different levels of granularity.
The first level concerns literals (\textit{i.e.,} numerical values, strings, etc.), for example the granularity of a product description as depicted in Figure~\ref{fig:knowledge-fusion-galaxy}.
The second level concerns concepts \textit{e.g.,} Korea \textit{vs.} Asia to indicate the location of Suwon-si as depicted in Figure~\ref{fig:taxonomy}.
On top of that, different scales of knowledge are possible as illustrated in Figure~\ref{fig:knowledge-fusion-galaxy} among triples extracted from $\mathcal{S}_{1}$ we have $\left\langle \text{Galaxy S23, hasPrice, ``800\euro''} \right\rangle$ and from source $\mathcal{S}_{2}$ we have $\left\langle \text{Galaxy S23, hasPrice, ``expensive''} \right\rangle$, the terms ``800\euro'' and ``expensive'' are indeed not in the same scale, one indicates the exact price while the second one gives a qualification of the price.
In all cases, the first and second sources provide complementary pieces of information for other attributes.
Figure~\ref{fig:knowledge-fusion-galaxy-2} shows the resulting graph after the reconciliation step that includes the knowledge alignment and the fusion step where the most complete representation of the Galaxy S23 entity is produced.

\begin{figure}[h]
	\centering
	\includegraphics[scale=0.95]{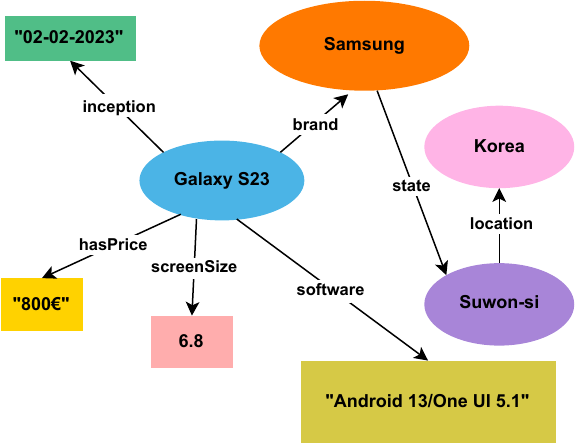}
	\caption{Resulting KG after reconciliation step (alignment and fusion).}
	\label{fig:knowledge-fusion-galaxy-2}
\end{figure}

In the next section, we survey several methods that address knowledge fusion for truth discovery.
We will use the terms ``truth inferring'', ``truth finding'' or ``truth discovery'' in the same way.

\subsection{Fusion Approaches}
\label{subsubsection:fusion-approaches}

Fusion methods are identified in Table~\ref{tab:fusion-recap}, which provides an overview and indicates if the methods can handle some characteristics of data or data sources such as numerical data, categorical data, data granularity, and dependency between data sources.

\begin{sidewaystable*}
	\begin{adjustbox}{width={\textwidth},totalheight={1\textheight},keepaspectratio}
		\begin{tabular}{lccccccclc}
			\toprule
            \multicolumn{1}{c}{\multirow{2}{*}{Model}} & \multicolumn{2}{c}{Task} & \multicolumn{1}{c}{\multirow{2}{*}{Modeling}} & \multicolumn{4}{c}{Awareness} & \multicolumn{1}{c}{\multirow{2}{*}{Datasets}} & \multicolumn{1}{c}{\multirow{2}{*}{Year}} \\
            \cmidrule{2-3}
            \cmidrule{5-8}
			\multicolumn{1}{c}{} & \multicolumn{1}{c}{TI} & \multicolumn{1}{c}{SQ} & \multicolumn{1}{c}{} & \multicolumn{1}{c}{Numerical Data} & \multicolumn{1}{c}{Categorical Data} & \multicolumn{1}{c}{Granularity} & \multicolumn{1}{c}{Source dependence} & \multicolumn{1}{c}{} & \multicolumn{1}{c}{} \\
            \midrule
            TruthFinder~\cite{yinHan2007} & $\bullet$ & $\bullet$ & Probabilistic & $\bullet$ & $\bullet$ & Vagueness & $\bullet$ & Book authors & 2007 \\
            \\
            ACCU~\cite{DongBerti2009} & $\bullet$ & $\bullet$ & Probabilistic & $\bullet$ & $\bullet$ & Vagueness & $\bullet$ & Synthetic & 2009 \\
            \\
            LFC~\cite{raykar2010} & $\bullet$ & $\bullet$ & Probabilistic & $\bullet$ & $\bullet$ & $\circ$ & $\circ$ & Digital mammography and Breast MRI & 2010 \\
            \\
            GTM~\cite{zhaoHan2012} & $\bullet$ & $\bullet$ & Probabilistic & $\bullet$ & $\circ$ & Fuzziness & $\circ$ & Wikipedia edit history of city population and People biographies & 2012 \\
            \\
            LTM~\cite{zhaoRubinstein2012} & $\bullet$ & $\bullet$ & Probabilistic & $\bullet$ & $\bullet$ & Completeness & $\circ$ & Book author, Movie director, and Synthetic & 2012 \\
            \\
            POPACCU~\cite{dongSaha2012} & $\bullet$ & $\bullet$ & Probabilistic & $\bullet$ & $\bullet$ & $\circ$ & $\circ$ & Books (from AbeBooks.com) and Flight & 2012 \\
            \\
            LCA~\cite{pasternack2013} & $\bullet$ & $\bullet$ & Probabilistic & $\bullet$ & $\bullet$ & $\circ$ & $\circ$ & Books, Population, Stocks, and FantasySCOTUS & 2013 \\
            \\
            CRH~\cite{LiLi2014} & $\bullet$ & $\bullet$ & Optimization problem & $\bullet$ & $\bullet$ & Vagueness & $\circ$ & Weather Forecast, Stocks, and Flight & 2014 \\
            \\
            CATD~\cite{LiLiCATD2014} & $\bullet$ & $\bullet$ & Optimization & $\bullet$ & $\bullet$ & $\circ$ & $\circ$ & City Population, Biography, and Indoor Floorplan & 2014 \\
            \\
            KBT~\cite{Dong2015} & $\bullet$ & $\bullet$ & Probabilistic & $\bullet$ & $\bullet$ & $\circ$ & $\circ$ & Triples collected by Knowledge Vault & 2015 \\
            & & & & & & & & Synthetic \\
            \\
            FaitCrowd~\cite{maLi2015} & $\bullet$ & $\bullet$ & Probabilistic & $\bullet$ & $\bullet$ & $\circ$ & $\circ$ & SFV and Dataset from crowdsourcing platform & 2015 \\
            \\
            DOCS\cite{zheng2016} & $\bullet$ & $\bullet$ & Probabilistic & $\bullet$ & $\bullet$ & $\circ$ & $\circ$ & ItemCompare, 4-Domain, Yahoo QA, and SFV & 2016 \\
            \\
            ASUMS~\cite{berettaHaripse2016} & $\bullet$ & $\bullet$ & Belief functions & $\circ$ & $\bullet$ & Fuzziness & $\circ$ & Synthetic and People biographies & 2016 \\
            \\
            KDEm~\cite{wan2016} & $\bullet$ & $\bullet$ & Optimization & $\bullet$ & $\circ$ & Completeness & $\circ$ & Synthetic and Population & 2016 \\
            \\
            SLiMFAST~\cite{Rekatsinas2017} & $\bullet$ & $\bullet$ & Probabilistic & $\bullet$ & $\bullet$ & $\circ$ & $\bullet$ & Stocks, Demonstrations, Crowds, and Genomics & 2017 \\
            \\
            MDC~\cite{liDu2017} & $\bullet$ & $\bullet$ & Representation Learning & $\circ$ & $\bullet$ & $\circ$ & $\circ$ & Dataset created from baobaozhido (crowdsourcing platform) & 2017 \\
            & & & Optimization problem & & & & & \\
            \\
            HYBRID~\cite{liDong2017} & $\bullet$ & $\bullet$ & Probabilistic & $\bullet$ & $\bullet$ & Completeness & $\circ$ & Book and Synthetic & 2017 \\
            \\
            TDH~\cite{jungKim2019} & $\bullet$ & $\bullet$ & Probabilistic & $\bullet$ & $\bullet$ & Fuzziness & $\bullet$ & BirthPlaces and Heritages & 2019 \\
            \\
            Record Fusion~\cite{Heidari2020} & $\bullet$ & $\circ$ & Softmax Classifier & $\bullet$ & $\bullet$ & $\circ$ & $\circ$ & Flight, Stock (1 \& 2), Weather, and Address & 2020 \\
            \\
            OKELE~\cite{cao2020} & $\bullet$ & $\bullet$ & Probabilistic & $\bullet$ & $\bullet$ & Completeness & $\circ$ & Synthetic & 2020 \\
            \\
            TKGC~\cite{huang2022} & $\bullet$ & $\bullet$ & Representation Learning & $\bullet$ & $\bullet$ & Completeness & $\circ$ & Dataset built from~\cite{cao2020} & 2022 \\
            & & & Probabilistic & & & & & Subgraph of Freebase as prior knowledge (KG) & \\
            \bottomrule
		\end{tabular}
	\end{adjustbox}
	\caption{Description of the tasks and characteristics of the data or sources that are ($\bullet$) or not ($\circ$) addressed by the knowledge fusion models. (TI) stands for Truth Inference, (SQ) stands for Source Quality.}
	\label{tab:fusion-recap}
\end{sidewaystable*}

\textbf{SLiMFAST}~\cite{Rekatsinas2017} leverages knowledge domain features to improve the quality estimation of data sources. 
For example, if the data is extracted from scientific articles, the authors suggest using features such as the number of citations to the article or the year of publication, which can influence the quality of the source.
Domain-specific features are included in the parameters of the logistic function that estimates source quality.
To merge the data, they apply statistical learning to estimate source quality, then apply probabilistic inference to predict true values.
To estimate the parameters, they use either the EM algorithm if the user does not provide labeled ground truth data, or the Empirical Risk Minimization algorithm if the user does.

\textbf{ACCU}~\cite{DongBerti2009} includes the interdependence between data sources in the truth discovery process.
The intuition is that it is possible that a single source could provide the true value and that all the other sources could provide false values knowing that some of them can copy on each other and therefore, spread false values. Thus, if a data source provides a value different from all the others, it is not systematically false.
They define the dependency between two sources if there is a part of their data that comes directly or transitively from a common source, and it is computed by Bayesian models.
Then, to discover the true value, they combine this dependency evaluation and the accuracy of the data sources which are computed in relation to the confidences of the values and dependencies between sources.

\textbf{POPACCU}~\cite{dongSaha2012}, unlike ACCU, considers that the data sources are independent and that only one value can be correct.

\textbf{CRH}~\cite{LiLi2014} (Conflict Resolution on Heterogeneous Data) estimates the reliability of a source, it uses all types of data simultaneously instead of focusing on a single one.
To do this, the authors use a loss function that measures the distance between the unknown truth and the values claimed by the sources for each type of data.
To initiate the source reliability score, it first applies a simple conflict resolution method, such as majority or average voting and models the estimation of source and truths weights as an optimization problem.

\textbf{MDC}~\cite{liDu2017} takes into account the semantic aspect of values. 
To illustrate the importance of semantics, in the paper the authors provide the following example: a first data source provides the true value ``common cold'', another source claims the value is ``sinus infection'' while a last source claims the value is ``bone fracture'', instead of examining all values at the same level, MDC calculates semantic proximity among values. 
This semantic proximity calculation allows evaluating how close a value is to the true value. 
The semantics of the values are captured by their vector representations learned by following the idea that if two values share similar words, then their vectors should be similar.

\textbf{DOCS}~\cite{zheng2016} is a system deployed on the Amazon Mechanical Turk that takes into account the precision of the answers of each worker to assign a specific task from a specific domain to the right worker.
Regarding true value inference, the system takes advantage of the inherent relationships between the reliability of workers (which can be seen as data sources) and the true value.
Thus, it considers two events: \textit{(1)} let $v$ be the value of an entity provided by a source $s$, if the quality of the values provided by $s$ of entities in the same domain as $v$ is high then $v$ is likely to be correct;
\textit{(2)} then, if a source $s$ often provides correct values for a domain $d$, then $s$ has high reliability for domain $d$.

\textbf{TruthFinder}~\cite{yinHan2007} claims that a fact is more likely to be true if it is provided by a reliable source, and that a source is reliable if it provides proven facts.
Thus, an interdependence between facts and sources appears and consequently TruthFinder uses three elements for its iterative trust discovery process: the trustworthiness of sources, the confidence of facts, and the influences between facts.
The trustworthiness $t(w)$ of a source $w$ is computed by
$$t(w) = \frac{\sum_{f\in F(w))}^{}s(\mathrm{t})}{\left| F(w) \right|},$$ and the confidence $s(\mathrm{t})$ of a fact $\mathrm{t}$ is computed by $$s(\mathrm{t})=1-\prod_{w\in W(\mathrm{t})}(1-t(w)).$$
Then, they use the logarithm to facilitate the computation and to obtain the final scores.
Although simple, these definitions integrate the influence of facts where values of different vagueness levels (\textit{e.g.,} ``Joe Biden'', ``J. Biden'', and ``Biden'') can increase the confidence of a fact and take into account the dependence between data sources by a dampening factor acting on the computation of the confidence score.

While most existing methods consider more generic values of a correct one as false, \textbf{TDH}~\cite{jungKim2019} leverages the hierarchical structure of knowledge (\textit{i.e.,} its one aspect of the granularity) to apply the fusion of data extracted from different sources.
The idea behind is that multiple values in the hierarchy of an entity could be correct even if the predicate is functional.
For example, in Figure~\ref{fig:taxonomy}, ``Korea'', and ``Asia'' are correct values for the location of ``Suwon'' even though one of both values is more specific.
Such a modeling should not negatively impact the assessment of the reliability of the source.
Instead of evaluating the value as $correct$ or $incorrect$, they consider three classifications, namely \textit{exactly correct}, \textit{hierarchically correct}, and \textit{incorrect}.
Therefore, each source has its own generalization tendency and reliability.

In a same way, \textbf{ASUMS}~\cite{berettaHaripse2016} adapts existing truth discovery models by considering that not all values are necessarily conflicting and identifies a partial order between the values of an attribute using the ``subClassOf'' and ``partOf'' relationships.
To do this, they use belief functions capable of modeling ignorance and uncertainty and allowing the incorporation of knowledge about the relations between values.
Thus, in this modeling several true values can coexist but of different granularity.
Therefore, all facts more generic than a certain fact are true and conflicting facts are facts located at the same hierarchical level but with different values.

\begin{figure}[h]
	\centering
	\includegraphics[scale=0.8]{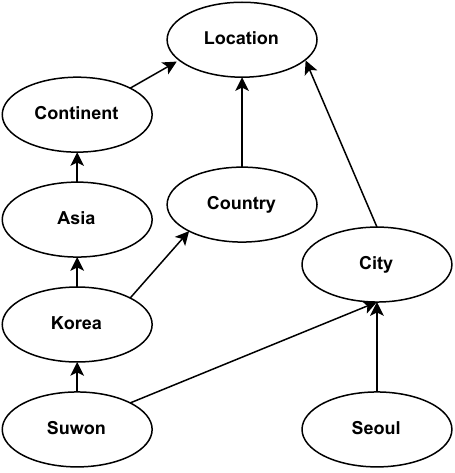}
	\caption{Illustration of a partial ordering.}
	\label{fig:taxonomy}
\end{figure}

\textbf{LFC}~\cite{raykar2010} also measures the performance of each source (\textit{e.g.,} annotators in the paper) by their specificity and sensitivity relating to the unknown gold standard dataset to give higher weights to the best-performing sources.
The estimations are iteratively performed via the EM algorithm, where the missing data is the true value initialized by applying a majority voting.
Specificity and sensitivity are then estimated in turn, and so on until convergence is reached.

\textbf{LCA}~\cite{pasternack2013} includes four models with different sophistication, namely: SimpleLCA, GuessLCA, MistakeLCA, LieLCA.
LCA is a probabilistic model where the true value is a multinomial latent variable.
To infer the truth, an EM algorithm is used to compute the trustworthiness of each source with respect to the claims it makes, then the true value is computed based on the trustworthiness of the sources.

\textbf{KBT}~\cite{Dong2015} focuses on assessing the quality of Web sources from which facts are extracted and also evaluated.
Facts are extracted as triples (subject, predicate, object) from Web pages using Knowledge Vault, which is composed of 16 different extractors.
The authors extend the ACCU model by improving the estimation of the reliability of a source by distinguishing the error that comes from the fact against the error that comes from the extraction of the fact.
However, they do not consider granularity or the fact that several correct values may coexist for a data item.

While other approaches are focused on categorical data, \textbf{GTM}~\cite{zhaoHan2012} (Gaussian Truth Model) tackles the truth finding task on numerical data.
This model focuses on the position of numerical claim values $v_{c}$ relative to others in terms of distance to find the truth.
To embed the notion of distance in their model, the authors consider the truth of each entity as a random variable and use it as the mean parameter in the probabilistic distribution for each claimed value of the observed entities.
To do this, they leverage a Gaussian distribution for its ability to model errors thanks to its quadratic penalty.
Regarding the evaluation of the quality of data sources, they assume that quality is related to the closeness of the claims to the truth.
Therefore, the quality of a source is modeled by the variance of the Gaussian distribution, \textit{e.g.,} a high-quality source is represented by a low variance.
As aforementioned, the model can take as input the output of another truth finding method or a basic truth estimate (\textit{e.g.,} mean or median value) to limit the effect of possible outliers on the maximum likelihood estimate (MLE).
The quality of each source is determined from a prior inverse Gamma distribution, and the truth for an entity is determined from a prior Gaussian distribution.
Then, to compute the truth and source quality, they perform an EM algorithm.

In the same manner as GTM, \textbf{LTM}~\cite{zhaoRubinstein2012} incorporates prior knowledge about sources or truth into the truth finding process and introduces the notion of two-sided source quality.
It simultaneously deduces the quality of the source and the truth, as both influence each other.
To compute the quality of data sources, the authors consider each source as a classifier, with its own confusion matrix.
Thus, the quality of a source is defined by its sensitivity (or recall) which corresponds to the ``false negative rate'' and by its specificity which corresponds to the ``false positive rate'' that are two independent measures.
Both sensitivity and specificity are generated from a Beta distribution: the parameters for specificity are ``the prior false positive count'' and ``the prior true negative count'' and the parameters for sensitivity are ``the prior true positive count'' and ``the prior false negative count''.
The prior truth probability is also modeled by a Beta distribution with the parameters ``the prior true count'' and ``the prior false count'' for each distinct (entity, value) pair.
The truth value is generated by a Bernoulli distribution with a parameter $\theta$, corresponding to the prior probability that the value is true.
Finally, the truth and the quality of sources are inferred by a Collapsed Gibbs Sampling.

\textbf{Record Fusion}~\cite{Heidari2020} merges knowledge by relying on integrity constraints, quantitative statistics, and provenance information if this latter is available.
To find the true value of each table cell, they use one classifier per column (attribute) present in the table (dataset). 
These softmax classifiers can be modular, for example a logistic regression, a decision tree, a neural network, and so on.
Three representation models are explored to create the feature vector, which will then be provided to the classifier. 
The first representation is the role of a cell at the column level, where three different strategies are proposed: the first acts on the format of the data (\textit{e.g.,} the letters of the alphabet are replaced by the token ``A'', numbers by ``N'' and characters (\textit{e.g.,} ```'', ``.'', ``''', or space) by ``S'' and they get a vector from a n-gram model), the second strategy is to cluster attribute values, then the last strategy consider a matrix of embeddings which will map all the values of a column (attribute) in a Euclidean space and for each cell, computes the distance between its position in space and the average position of the other values. 
The second concerns the role of the cell at the row level (tuple), \textit{i.e.,} it captures the relationship of the attribute with other attributes in its row (entity).
Two signals are leveraged, the first one includes the counts of pairs of attributes that are seen together, and the second one captures how often a cell occurs among rows within its own entity. 
Finally, the third concerns the role of the cell in relation to the complete dataset (table), \textit{i.e.,} takes into consideration the number of denial constraint violations, includes the source information only if available since entities can have different provenance and each source can have different levels of trust. 
Then, the last step consists in training the different classifiers by a stage-wise additive model for a number of $T$ iterations: \textit{(1)} they learn the softmax classifiers with the original dataset, \textit{(2)} use previous predictions to construct a new dynamic feature, \textit{(3)} and learn again the classifiers using these new sets of features. 
For cells where the label is unknown, they assign a majority vote as their weak labels. 
They obtain good performance on datasets Flight, Stock, Weather, and Address about 94\%-98\% of precision.

In contrast to many fusion truth inferring approaches, \textbf{FaitCrowd}~\cite{maLi2015} measures the quality of data sources over several degrees, with one quality degree for each knowledge topic for a crowdsourcing case.
FaitCrowd represents the expertise of each source for each topic by a Gaussian distribution.
Then, it models the true value provided by a source for a certain question on a given topic as a logistic function depending on the contribution ratio of the source on the topic, the expertise of the source and a bias.
To estimate the parameters, the model uses the Gibbs-EM inference method that alternates between Gibbs sampling and gradient descent.

\textbf{TKGC}~\cite{huang2022} takes advantage of prior knowledge from the KG when feeding it and considers that the noise affecting the truth is represented by a probability distribution determined by the data source.
To estimate the difference between the truth value and the value supplied by a source, the authors use a difference function adapted to each data type, \textit{i.e.,} categorical data, numerical or datetime value, and string.
This difference function takes as input the representation vectors previously learned in a fact scoring setting for KG completion.
In fact, the probability of this latter function follows a Gaussian distribution $\mathcal{N}(0, (k_{a}\sigma_{s}^{2})$ where $k_{a}$ is a regularization factor and $\sigma_{s}$ representing the noise of a data source.
Then the truth inference is performed through a semi-supervised algorithm.

\textbf{OKELE}~\cite{cao2020} models the probability of a fact being true by a latent random variable following a Beta$(\beta_{0}, \beta_{1})$ distribution, with $\beta_{1}$ corresponding to prior true count of the fact and $\beta_{0}$ corresponding to its prior false count.
The quality of a data source is represented by its error variance $\omega_{s}$ that follows a scaled inverse chi-squared distribution Scale-inv-$\chi^{2}(v_{s}, \tau_{s}^{2})$ representing the number of facts provided by the source $v_{s}$ with variance $\tau_{s}^{2}$.
The authors argue that this distribution handles the effect of dataset size in the case of long-tail entities.
The truth inference is performed by leveraging prior knowledge from existing KGs to identify whether an attribute expects a single or multiple values.

\textbf{HYBRID}~\cite{liDong2017} tackles the TI task about knowledge on tail verticals and experiment the fusion by considering two assumptions: single-truth and multi-truth.
Before applying data fusion, they collect ``evidences'' on entities by looking for whether a source contains the subject and object of the original triple.
To do this, they use three types of sources: knowledge bases (Freebase and Knowledge Vault), the Web, and query logs.
The provenance information such as the URL where the system found the evidence and the pattern are retained.
Once the evidence retrieval is complete, HYBRID leverages the number of truths for each type of data items as a prior probability (for example, a mobile phone has only one year of creation, or we could consider between two and height buttons).
Therefore, when a single true value or multiple true values are expected, it applies a single-truth model or a multi-truth model respectively.
To assess the quality of data sources, two metrics are used: \textit{precision}, \textit{i.e.,} the probability when a source provides a value, a truth exists and \textit{recall}, \textit{i.e.,} when a truth exists, the source provides a value.

\textbf{CATD}~\cite{LiLiCATD2014} addresses the problem of fusion in a context where data sources provide a few claims and where estimating their quality is difficult due to the lack of data.
Quality of sources are estimated by a Gaussian distribution whose mean represented the bias of the source, \textit{i.e.,} its intentional behavior to provide false information and variance represents the reliability degree of the source.
To cope with the problem of a small amount of data available from a source, they consider a confidence interval of the variance to represent their reliability.
Finally, CATD applies an optimization algorithm by initializing the true values with a simple method (\textit{e.g.,} a median of the values) and starts by estimating the quality of the sources, which depends on their claimed values, then estimates the true values.

\textbf{KDEm}~\cite{wan2016} replaces the concept of true value with the concept of trustworthy opinions on the value of an entity. 
This model allows several true values for an entity's attribute, and consequently considers a form of knowledge granularity. 
KDEm leverages the kernel density estimation with a Gaussian kernel and extends it by adding the weights of the sources to estimate the probability distributions of values for each attribute of an entity.
To find true values it combines the density estimation with a threshold and detects outliers that are below this threshold.

\section{Uncertainty Representation}
\label{section:uncertainty-representation}

Handling knowledge uncertainty throughout the data integration process also includes its representation in the KG.
Uncertainty can be represented on different value scales such as numeric, alphanumeric, textual, or intervals of values.
If these different levels of uncertainty are used for reconciliation, they need to be preserved and represented in the KG as metadata in order to retain a history and could possibly be useful for resolving future conflicts~\cite{Ilyas2022}.
The inclusion of uncertainty in the KG also enables the selection of knowledge in relation to their confidence and contributes to maintain the quality of the graph~\cite{Weikum2021}.
Several works deal with querying UKGs.
For example, in~\cite{hartig2009}, Hartig presents tSPARQL that extends RDF model and its query language SPARQL to handle uncertainty.
In \cite{dellal2019}, the authors solve the failing RDF query problem, \textit{i.e.,} when a user obtains an empty answer, that can arise when a user queries the graph with a high confidence threshold.
To do this, they use tSPARQL~\cite{hartig2009} and propose answers obtained by Minimal Failing Subquery, \textit{i.e.,} the minimal subquery contained in the failed main query, and Maximal Succeeding Subquery, \textit{i.e.,} the maximal subquery that succeeded under the confidence threshold provided by the user.
In~\cite{meiser2011}, the authors propose a reasoner called URDF that solves data uncertainty for SPARQL queries.
Another work tackles the task of UKG querying using UKG embeddings~\cite{fei2024}.
Therefore, it is important to choose the best knowledge representation when building a KG according to few criteria described in Section~\ref{subsubscetion:uncertainty-data-model}.
For instance, several sources may provide the same data, hence the model must also be able to include all provenance information (\textit{i.e.,} mutliple data sources).
For specific applications, some information to assign to triples could be required \textit{e.g.,} provenance information, the uncertainty from extraction algorithms or other, or spatial and temporal information~\cite{nguyenBodenreider2014, dividinoSizov2009, hartig2017, carroll2005, orlandi2021}.
Some formalisms of the Semantic Web offer possibilities for representing this uncertainty through metadata.
Metadata is data about data defined within the RDF model and is important to estimate the validity of the information~\cite{dividino2009}.
We present the uncertainty representation at the ontology level in Section~\ref{subsubsection:uncertainty-ontology} and at the data model level in Section~\ref{subsubscetion:uncertainty-data-model}.

\subsection{Uncertainty representation at ontology level}
\label{subsubsection:uncertainty-ontology}

An ontology entails three notions, namely conceptualization, explicit and formal specification, and sharing~\cite{guarino2009, gruber1993}.
The conceptualization is an abstract view of a domain, including the relevant concepts and entities, and relates them together.
In this way, ontologies make domain knowledge understandable by the machine and enable reasoning about knowledge by defining rules, constraints, and the domain and range of relations~\cite{yang2005}. 
\cite{anand2022} provides a table that summarizes the usual components that form an ontology (refer to this table for further details).
OWL (Ontology Web Language) enables to describe an ontology of a knowledge domain through individuals, classes or concepts, and properties. 
It is based on description logic and is part of the W3C's recommendations.
Despite its ability to define rich ontologies, OWL cannot natively represent and reason about uncertainty since this latter is based on crisp logic, \textit{i.e.,} a statement is true or false contrary to fuzzy logic for example \cite{zhongli2006}.
Thus, in \cite{costa2008} the authors argue that the lack of ways for handling uncertain information affects the requirements of the Semantic Web.
To model and handle uncertainty in an ontology by including their uncertainty theory, most approaches extend the OWL ontology~\cite{mcguinness2004}.
These uncertainty theories include the probability theory \textit{e.g.,} Bayesian Network (BN), Fuzzy logic, Belief functions \textit{e.g.,} Dempster-Shafer (DS) theory.

\textbf{Fuzzy-OWL}~\cite{stoilos2005} extends OWL with fuzzy sets theory for covering vague knowledge.

\textbf{OntoBayes}~\cite{yang2005} integrates BNs into OWL in order to preserve the advantages of both.
Three OWL classes are introduced: \textit{PriorProb}, \textit{CondProb}, and \textit{FullProbDist} of type \textit{ProbValue} (value between 0 and 1) to manage probabilities.

\textbf{PR-OWL}~\cite{costa2008} aims to provide a probabilistic extension of OWL since the probability theory can represent uncertainty by combining Bayesian probability theory with First Order Logic.
In addition to OWL, the ontology includes the statistical regularities that characterize the knowledge domain, the knowledge that is incomplete, inconclusive, ambiguous, unreliable and dissonant, then the uncertainty associated with this knowledge.
It has the ability to perform probabilistic reasoning with incomplete or uncertain information conveyed through an ontology but requires RDF Reification (presented in Section~\ref{subsubscetion:uncertainty-data-model}) since a probabislitic model includes more than on individual (N-ary relations).

\textbf{BayesOWL}~\cite{zhongli2006} completes OWL for representing and reasoning with uncertainty based on Bayesian Networks.
The BayesOWl model includes a set of structural translation rules to convert an OWL ontology into a directed acyclic graph of a Bayesian Network.
It provides the encoding of two types of probability, namely priors and pair-wise conditionals through two defined OWL classes \textit{PriorProb} and \textit{CondProb}.
Thus, a prior probability for a concept is defined as an instance of class \textit{PriorProb} with two properties named \textit{hasVariable} and \textit{hasProbValue}.
Then a conditional probability is represented through an instance of class \textit{CondProb} with the same properties as the above instance and a property \textit{hasCondition}.

\textbf{URW3-XG}~\cite{laskey2008} provides an ontology as a starting point to be refined.
A sentence about the world is asserted by an agent.
The uncertainty of a sentence has a relation \textit{hasUncertainty} with a \textit{derivationType}, \textit{uncertaintyType}, \textit{UncertaintyModel}, and a \textit{nature}.
\textit{UncertaintyType} includes the ambiguity, empirical uncertainty, randomness, vagueness, inconsistency and incompleteness.
\textit{UncertaintyModel} includes probability, fuzzy logic, belief functions, rough sets, and other mathematical models for reasoning under uncertainty.
And \textit{UncertaintyNature} is whether aleatoric, \textit{i.e.,} ontic or epistemic.

\textbf{Poss-OWL 2}~\cite{safia2014} provides an extension of OWL 2 to represent incomplete and uncertain knowledge with a possibilistic viewpoint.
The ontology has three main classes: \textit{concept}, \textit{role}, and \textit{axiom}.
\textit{Concept} is the equivalent of concept constructor of OWL 2 with an added degree that stands for the certainty level of the concept.
\textit{Role} represents the properties of objects and data.
Then, \textit{Axiom} corresponds to the possibilistic axioms (PossTBoxAxiom and PossABoxAxiom) where each axiom is associated to a real value representing the certainty level of the axiom.
The main issue of Poss-OWL 2 is that it only focus on the description of uncertainty at class-level.

\textbf{Riali \textit{et al.}}~\cite{riali2019} proposes a probabilistic extension of fuzzy ontologies in order to model vague, imprecise, probabilistic knowledge since fuzzy OWL only models vagueness.
Riali \textit{et al.} also provide a comparison of different approaches for modeling uncertainty in an ontology such as PODM \cite{hlel2016}, HyProb-Ontology \cite{abdul2016}, etc.

\textbf{mUnc}~\cite{djebri2019} aims to unify the different uncertainty theories within a single ontology.
The ontology includes the following theories: probability, evidence of Dempster-Shafer, and possibility theory.
mUnc allows publishing uncertainty theories alongside their features and computation methods.
Each uncertainty theory is linked to a set of features and operators.
The features correspond to the metrics on which uncertainty theory is based to indicate the degree of truth, credibility, or the likelihood of a sentence.

\subsection{Uncertainty representation at data model level}
\label{subsubscetion:uncertainty-data-model}

The basic RDF model cannot natively inject values directly into the edges.
On the other hand, there are ways and other graph representations to circumvent this limitation.
To detail these graph representations, we consider that we want to represent the uncertainty through a confidence score $s \in \left[ 0,1 \right]$, with 0 representing low confidence and 1 representing high confidence.
In~\cite{angles2022}, the authors use 10 criteria to compare five data representation models: RDF, RDF*, Named Graph, Property Graph and their model Multilayer Graph.
Among these 10 criteria, we consider that two main criteria are required to represent the confidence score and the provenance of a RDF triple.
The first criterion is \textit{edge annotation} and refers to the ability of the representation model to assign attribute-value pairs to an edge.
The second one is \textit{edge as nodes}, meaning that an edge can be referenced as multiple nodes.
Therefore, we review the data representation models \textit{w.r.t.} these two criteria and their pros and cons. 
We illustrate the models with the triple <JoeBiden, isPresident, UnitedStates> associated with the confidence score ``0.911'' in Figure~\ref{fig:uncertainty-representations}.
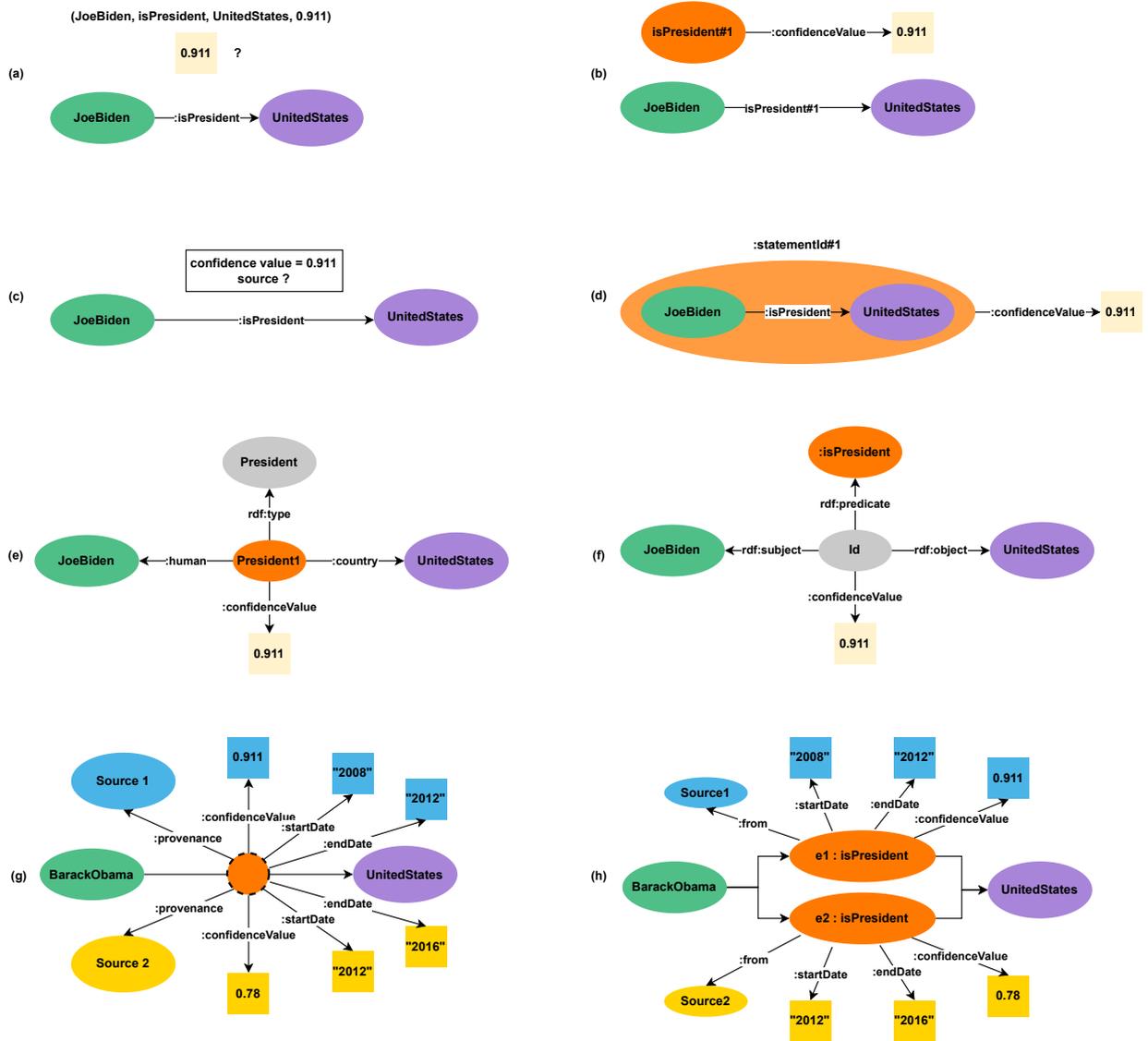
\begin{figure*}[h]
	\centering
	\includegraphics[width=\textwidth]{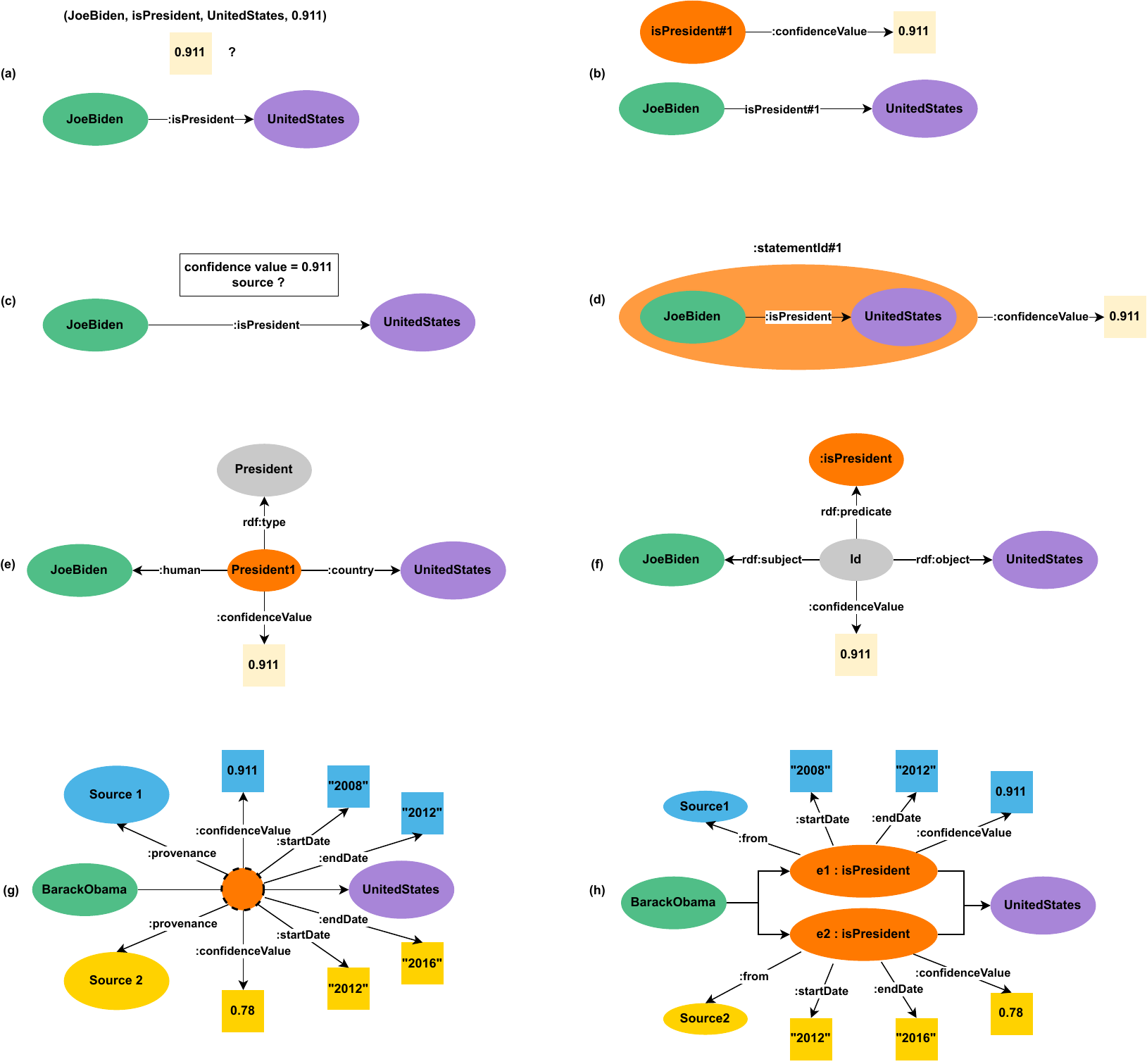}
	\caption[]{Illustration of different ways to embed the uncertainty (captured by  a confidence score) of a triple in the KG representation: (a) RDF, (b) Singleton Property, (c) Property Graph, (d) Named Graph, (e) N-ary, (f) RDF Reification, (g) RDF-star (RDF*), (h) Multilayer Graph. \raisebox{-2pt}{\begin{tikzpicture}[scale=0.2]
        \node[draw=black, circle, dashed, fill=orange]{} ;
    \end{tikzpicture}} in RDF-star illustration depicts the triple <BarackObama, isPresident, UnitedStates>.}
	\label{fig:uncertainty-representations}
\end{figure*}

\textbf{Singleton Property}~\cite{nguyenBodenreider2014} uses a new type of property called ``singleton property'', which corresponds to a unique property linked to an URI between two entities.
This unique property can be used as a node to which additional relations can be added.
For example, the singleton property in Figure~\ref{fig:uncertainty-representations}(b) is ``isPresident\#1'', then this node is used to add the confidence value ``0.911''.
Despite the fact that the singleton property is convenient for a compact meta-level representation, this modeling introduces many unique predicates and affects data querying~\cite{rupp2022, hartig2017}.

\textbf{Property Graph} puts additional information about triples that are stored as a list of key/value pairs at edges in the graph.
For example, in Figure~\ref{fig:uncertainty-representations}(c), the confidence value and the provenance are attached to the relation ``:isPresident''.

\textbf{Named Graph}~\cite{hernandezHogan2015} extends the RDF triple model and allows to indicate a triple as a subgraph denoted by an IRI.
This subgraph with an identifier can be used to add meta-information.
In Figure~\ref{fig:uncertainty-representations}(d), the original RDF triple <JoeBiden, :isPresident, UnitedStates> is identified by ``:statementId\#1'', which is used as the subject in the triple <:statementId\#1, :confidenceValue, 0.911>.
This modeling, which corresponds to nested graphs, is well-supported in the SPARQL standard and well-suited for representing provenance data~\cite{rupp2022}.

\textbf{N-ary}~\cite{manola2004} creates a node to represent a relation concept whose triples linked to this node correspond to the arguments of the relation.
For example, in Figure~\ref{fig:uncertainty-representations}(e), an intermediate node ``President1'' of type ``President'' characterizes the relation ``:isPresident'', then meta-data can annotate the relation.
The main drawback of this representation is its cumbersome syntax, which increases the complexity of the KG since the n-ary relation must be divided into several binary relations.

\textbf{RDF reification}~\cite{hayes2014} consists in creating an Internationalized Resource Identifier (IRI) or blank node that plays the role of the subject of all triples, as depicted in Figure~\ref{fig:uncertainty-representations}(f).
To represent the former triple it uses three new relations namely \textit{rdf:subject}, \textit{rdf:predicate}, and \textit{rdf:object} then as many relations as it needs to add metadata.
This modeling was the first way to make statements about statements~\cite{rupp2022}.
This method is simple, but its syntax is too verbose since each statement must be reified, which considerably increases the size of the KG, makes queries and RDF data exchange more complex~\cite{nguyenBodenreider2014, angles2022, rupp2022, hartig2014}.

\textbf{RDF-star}~\cite{hartig2017} is an extension of the RDF model proposed by the Semantic Web community.
A RDF triple is a tuple $t^{*} \in (T^{*} \times E) \times R \times (T^{*} \times E \times L) \in T^{*}$ and RDF-star triple is a tuple $t^{*} \in (T^{*} \times E) \times R \times (T^{*} \times E \times L) \in T^{*}$ where $E$ is the set of entities, $R$ is the set of relations, $L$ the set of literals, and $T^{*}$ is the set of RDF-star triples.
RDF-star can extend an existing RDF model expressively, since the metadata of triples are simply added as objects of them~\cite{kasenchak2021}.
For example, in Figure~\ref{fig:uncertainty-representations}(g), metadata such as provenance or confidence scores are added directly to the triple <BarackObama, :isPresident, UnitedStates> illustrated by \raisebox{-2pt}{\begin{tikzpicture}\node[draw=black, circle, dashed, fill=orange]{} ;\end{tikzpicture}}.
In addition to the ability to add metadata at the statement level without modifying the remaining data, RDF-star has its own query language called SPARQL-star which reduces compatibility issues~\cite{keskisarkka2020}.
A comparison on Wikidata have shown that RDF-star performs better than reification, n-ary, and named graph representations in terms of the number of triples, loading time, and storage capacity~\cite{kasenchak2021}.
However, RDF-star cannot consistently represent the same metadata with different values for the same triple~\cite{hartig2014, angles2022}.
Indeed, in Figure~\ref{fig:uncertainty-representations}(g), there are different start dates that refer to the same triple.

\textbf{Multilayer Graph}~\cite{angles2022} unifies the various advantages of the other representations we have described so far into a single, simple, and flexible (whether at node or statement level) model by introducing the notion of ``layer''.
To explain this, we use the notations and definition of a multilayer graph from~\cite{angles2022}: given $Obj$ a universe of objects that contains strings, numbers, IRIs and so on, a multilayer graph is defined as $G=(O, \gamma)$ where $O \subseteq Obj$ is a set of objects and $\gamma : O \to O \times O \times O$ is a partial mapping that models directed, labeled and identified edges between objects.
The layers in the multilayer graph come from the nested structure of edge ids.
The layer of an object $o \in O$, described as layer($o$) is defined as follows: if $o$ is not an edge id, then layer($o$) = 0; otherwise if $\gamma(o) = (n_{1}, l, n_{2})$ then layer($o$) = max\{layer($n_{1}$), layer($l$), layer($n_{2}$)\} + 1.
Figure~\ref{fig:layers-representation} depicts the layer representation of (h) from Figure~\ref{fig:uncertainty-representations}(h).
This data model enables to unambiguously represent multiple provenances and different confidence scores in a triple.

\begin{figure}[h]
	\centering
	\includegraphics[width=0.45\textwidth]{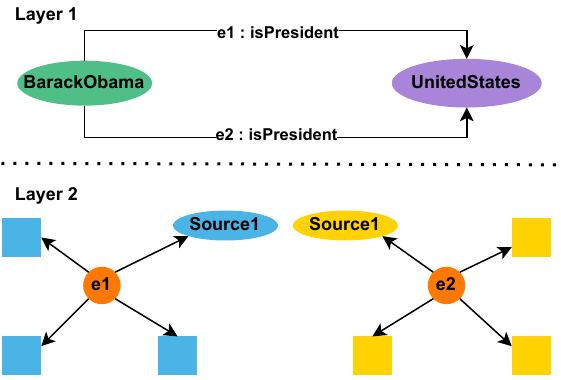}
	\caption{Illustration of the notion of ``layer'' where the first layer can be seen as a single triple <BarackObama, isPresident, UnitedStates> while the second layer contains different sets of information about the triple.}
	\label{fig:layers-representation}
\end{figure}

\section{Discussion and perspectives}
\label{section:perspectives}

Throughout this survey, we have seen that several approaches exist to represent uncertainty within KGs.
This is made possible by the development of ontologies that include multiple theories enabling uncertainty to be manipulated in addition to data models whose flexibility to include metadata about metadata enable additional information to be associated with extracted triples such as confidence scores.
However, we can argue that methods for integrating knowledge after its extraction, still overlook uncertainty in their modeling, despite the recently developed methods for embedding UKGs to perform link prediction, KG completion, or confidence prediction.
Taking into account the provenance information and the different levels of uncertainty operating at different locations in the knowledge integration pipeline, namely in knowledge (\textit{i.e.,} deltas), data sources, and all components of the pipeline (\textit{i.e.,} extraction, alignment, and fusion) would be beneficial to preserve the traceability, strengthens the quality of the KG, and enables graph querying by specifying a confidence level.

For the alignment task, the approaches do not take into account the uncertainty of the knowledge to be aligned and make the assumption that the knowledge are deterministic. 
On the contrary, many approaches that tackle KG completion tasks take knowledge uncertainty into account in their models.
We believe that extending these models to the task of knowledge alignment would be beneficial. 
For example, using embedding models of uncertain graphs for mapping-based alignment methods where a transformation function between the two embedding spaces of the two KGs to be aligned is learned. 
Or simply include confidence scores in neighbor aggregation for GNN-based models.

Once the knowledge has been aligned, it needs to be merged. 
We have seen several methods dealing with different knowledge characteristics such as granularity or numerical values.
Knowledge granularity is an essential aspect when building a KG from multiple heterogeneous sources. 
Indeed, if we leverage several popular data sources such as Wikipedia or Wikidata and one data source specific to the domain the KG is intended to represent, we are likely to face differences in granularity that we need to manage. 
If we use the simplest fusion approaches, such as majority voting or averaged voting, the graph will not contain the most specific knowledge.
We have seen that most fusion methods do not handle this aspect of granularity, and consider that only one true value exists. 
Only a few methods tackle this aspect by considering a partial order between the values to be fused or a semantic distance for categorical data. 
We therefore recommend developing this aspect further in the modeling of fusion models, for example by estimating a granularity score for a data source in parallel with its trustworthiness score, depending on the needs of KG builders.
One way of solving this problem is to further develop  fusion models to capture the correlation between the attributes of the entities and to identify any inconsistencies in one or more of its attributes.
Current fusion models incorporate a confidence score that embodies the trustworthiness of data sources to infer truth in knowledge fusion.
Nevertheless, the confidence in extraction algorithms and other components is not accounted for.
This modeling is not a problem when the same entities can be extracted from multiple sources.
However, when we deal with long-tail entities and when few data sources provide knowledge about them (for example, two data sources), if one contradicts the other and their confidence score are close, the fusion model may have difficulty to find the true value while other confidence scores such as in extraction could guide the fusion model.
We also advocate for better fusion models, since most proposals are specifically adapted to numerical data (which represents a relative small part of the entire data involved in KGs),  only deal with categorical data, or consider both types of data but handle them in the same way.
We believe that it would be more advantageous to consider the different data types in different ways, but within a single framework.

\section{Conclusion}
\label{section:conclusion}

In our current world, where knowledge may be noisy, contradictory  and  of different granularity, uncertainty should be taken into account when constructing a KG from multiple and heterogeneous data sources.
In fact, since KG construction  relies on automatic knowledge extractions, other levels of uncertainty should be accounted for.

In this paper, we proposed a classification of knowledge related uncertainty into two categories: uncertainty leading to contradictions and uncertainty leading to granularity disparities.
We then discussed a theoretical pipeline for the refinement of uncertain knowledge to be integrated in KG construction.
This pipeline consists of four main tasks: knowledge representation (including uncertainty and provenance in the KG), knowledge alignment, knowledge fusion, and consistency checking.
We also discussed challenges and perspectives on the integration of uncertain knowledge into a KG.

In particular, we have pointed out that tasks such as link prediction and KG completion are currently tackled with representational methods (embeddings) that take into account uncertainty.
Knowledge alignment is a well-studied topic, with a wide range of models available from rule-based models to deep learning models, for which we  provided a brief overview of existing methods. 
We also revisited knowledge fusion approaches, most of which based on probabilistic models, and estimated both the trustworthiness of data sources and true values.
However, knowledge integration remains a challenging topic for future research.
While the representation of uncertainty in a KG has received attention over the last few years (both at the ontological level and at the data model), the current knowledge integration approaches addressing both tasks remain limited in their scope (not taking into account all types of uncertainty and of knowledge deltas since they are only concerned with uncertainty).

\nocite{*}
\bibliographystyle{ios1}   
\bibliography{bibliography}      
\end{document}